\documentclass{article}
\usepackage[main,final]{neurips_2026}

\usepackage{amsmath,amssymb,amsfonts,amsthm,mathtools}

\usepackage[utf8]{inputenc}
\usepackage[T1]{fontenc}
\usepackage{hyperref}
\usepackage{url}
\usepackage{booktabs}
\usepackage{nicefrac}
\usepackage{microtype}
\usepackage{xcolor}
\usepackage{graphicx}
\usepackage{marvosym}
\usepackage{subcaption}
\usepackage{multirow}
\usepackage{float}
\usepackage{algorithm}
\usepackage{algorithmic}

\usepackage[capitalize,noabbrev]{cleveref}

\theoremstyle{plain}

\theoremstyle{definition}

\theoremstyle{remark}

\crefname{assumption}{Assumption}{Assumptions}
\Crefname{assumption}{Assumption}{Assumptions}

\title{DeformMaster: An Interactive Physics-Neural World Model for Deformable Objects from Videos}

\author{
  Can Li \\
  Nankai University
  \And
  Zhoujian Li \\
  Zhejiang University
  \And
  Ren Li \\
  Southern University of \\
  Science and Technology
  \And
  Jie Gu \\
  Rightly Robotics, A4x
  \AND
  Lei Lei \\
  Shanghai Innovation Institute, \\
  University of Science and Technology of China
  \And
  Jingmin Chen \\
  Rightly Robotics, A4x
  \And
  Lei Sun~\textsuperscript{\normalfont\Letter} \\
  Nankai University
}

\begin{document}

\maketitle
\begingroup
\renewcommand{\thefootnote}{}
\makeatletter
\long\def\@makefntext#1{\noindent#1}
\makeatother
\footnotetext{\Letter\ Corresponding author: Lei Sun (\texttt{sunl@nankai.edu.cn}). Can Li (\texttt{canli@mail.nankai.edu.cn}) and Lei Sun are with the College of Artificial Intelligence, Nankai University. Can Li, Zhoujian Li, and Lei Lei interned at Rightly Robotics, A4x.}
\endgroup

\begin{abstract}
World models for deformable objects should recover not only geometry and appearance, but also underlying physical dynamics, interaction grounding, and material behavior. Learning such a model from real videos is challenging because deformable linear, planar, and volumetric objects evolve under high-dimensional deformation, noisy interactions, and complex material response. The model must infer a physical state from visual observations, roll it forward under new interactions, and render the resulting dynamics with high visual fidelity. We therefore present DeformMaster, a video-derived interactive physics-neural world model that turns real interaction videos into an online interactive model of deformable objects within a unified dynamics-and-appearance framework. DeformMaster preserves structured physical rollout while using a neural residual to compensate for unmodeled effects, grounds sparse hand motion as distributed compliant actuators for hand-continuum interaction, represents material response with spatially varying constitutive experts, and drives high-fidelity 4D appearance from the predicted physical evolution. Experiments on real-world deformable-object sequences demonstrate DeformMaster's ability to roll out future dynamics and render dynamic appearance, outperforming state-of-the-art baselines while supporting novel action rollout, material-parameter variation, and dynamic novel-view synthesis. Code and video demos are available at the project page: {\definecolor{abstractlinkblueviolet}{HTML}{7366CC}\hypersetup{urlcolor=abstractlinkblueviolet}\color{abstractlinkblueviolet}\bfseries\itshape\urlstyle{same}\url{https://can-lee.github.io/deformmaster-web/}}.

\end{abstract}

\section{Introduction}

World models should encompass not only scene geometry, appearance, and temporal motion, but also the underlying physical attributes, governing dynamics, and causal interactions. Such models are especially important for embodied AI, where an agent must predict how the world will change under its own actions rather than merely reconstruct observations. Deformable objects make this goal particularly challenging: linear, planar, and volumetric objects evolve in high-dimensional state spaces, and their evolution is dictated by distributed strain, complex material response, self-contact, and external forces. A useful deformable-object world model must therefore infer the underlying physical state, support online interaction by rolling it forward under novel actions, and render its evolving appearance from novel views.

Existing methods have made some progress in reconstructing or generating dynamic deformable scenes. Neural and Gaussian representations \citep{mildenhall2020nerf,kerbl2023gaussian} can recover high-quality appearance from observations, and physics-aware reconstruction methods further fit physics engines for deformable objects from visual observations \citep{li2023pacnerf,cai2024gic,springgaus2024,jiang2025phystwin}. However, parameter identification within an idealized physics model can still struggle to account for real-world phenomena beyond the model assumptions. Recent works further exploit video diffusion either as a dynamic prior to supervise physics fitting \citep{zhang2024physdreamer,liu2025physflow} or as a generative engine for 4D synthesis \citep{lu2026phys4d}. However, video generation models \citep{yang2025cogvideox} primarily imagine how the world looks; they often lack a reliable understanding of action-conditioned dynamics, making them difficult to control through explicit interactions and prone to hallucinated dynamics that do not match the real physical scene.

Several lines of work attempt to introduce physical controllability, but each leaves a critical gap. Physical digital twins such as PhysTwin \citep{jiang2025phystwin} and Spring-Gaus \citep{springgaus2024} couple simulation substrates with Gaussian appearance, while EMPM \citep{chen2026empm} fits differentiable MPM for deformable object manipulation; yet fixed physics substrates and pure parameter fitting can hinder the model's ability to generalize to diverse real-world videos. Learned dynamics models based on particle-graph or particle-grid networks improve flexibility \citep{sanchez2020gns,zhang2025pgnd,zhang2024gsdynamics,zhang2024adaptigraph}. However, fully learned transitions often suffer from heavy data dependency, remain tied to the training distribution, and drift significantly during long-horizon novel-action rollouts. Hybrid physics-generative systems use physics to carry action semantics into 4D content \citep{li2025wonderplay,zhan2026perpetualwonder,liu2026realwonder}, but they primarily target generation rather than learning object dynamics from observed real-world interactions.

Our key insight is that real-world deformable-object world modeling should jointly recover underlying physical dynamics, interaction grounding, complex material behavior, and appearance tied to the evolving physical state. This requires more than fitting a simulator or learning visual motion alone: the model must remain stable under long-horizon interaction, absorb deviations from idealized physics, translate noisy observed contacts into effective actuation, and keep rendered appearance consistent with the predicted physical evolution. To address these requirements, we propose DeformMaster, an instance-specific interactive physics-neural world model that turns real interaction videos into an online queryable representation for novel action rollout, novel material-parameter variation, and dynamic novel-view synthesis, as illustrated in \cref{fig:teaser}. We summarize the contributions as follows:

\begin{figure}[t]
    \centering
    \includegraphics[width=\linewidth]{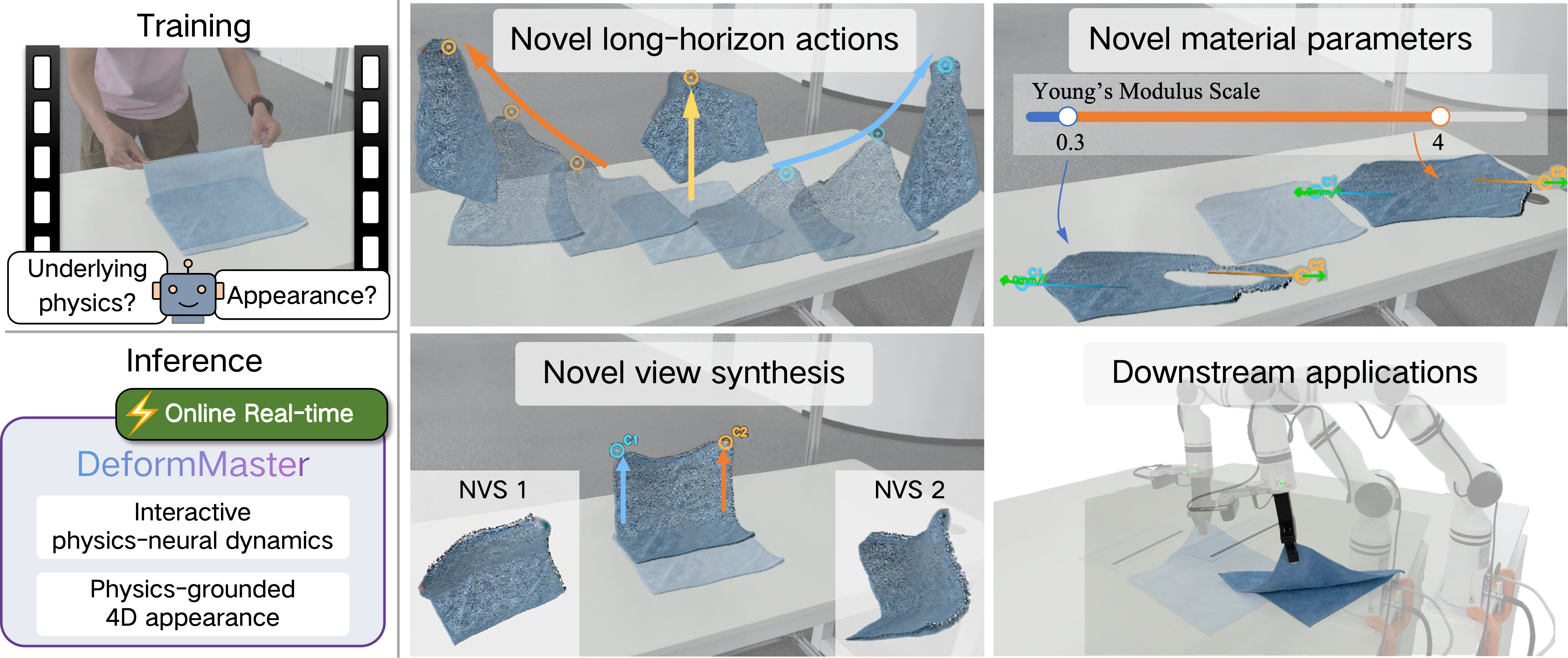}
    \caption{DeformMaster turns a phone-captured monocular video of deformable objects into an online interactive world model. By recovering both underlying physics and high-fidelity appearance, it supports long-horizon rollouts under novel actions, material-parameter variation, and novel-view synthesis, facilitating downstream embodied applications. ({\urlstyle{same}\url{https://can-lee.github.io/deformmaster-web/}})}
    \label{fig:teaser}
\end{figure}

\begin{enumerate}
    \item We propose Physics-Neural Particle-Grid Dynamics (PNPGD) that augments differentiable physics with a neural residual, preserving physics-guided rollouts while compensating for unmodeled real-world effects.
    \item We propose Distributed Compliant Actuators (DCA), which turn noisy sparse hand tracks into compliant, spatially distributed actuation for stable and effective hand-continuum interaction.
    \item We introduce a Mixture of Constitutive Experts (MoCE) that blends canonical material laws with spatially varying weights to capture heterogeneous material response.
    \item We develop DeformMaster, a video-derived deformable-object world model that pairs interactive physics-neural dynamics with physics-grounded high-fidelity 4D appearance.
\end{enumerate}

The rest of the paper is organized as follows. \Cref{sec:related} reviews related works. \Cref{sec:method} and \Cref{sec:experiments} present our proposed DeformMaster and its evaluation. \Cref{sec:conclusion} concludes the paper.

\section{Related Work}
\label{sec:related}

\paragraph{Reconstruction and physics simulation of deformable objects.}
Physics-based reconstruction methods aim to recover simulatable representations of deformable objects from visual observations. PAC-NeRF \citep{li2023pacnerf}, GIC \citep{cai2024gic}, PhysGaussian \citep{xie2024physgaussian}, OmniPhysGS \citep{lin2025omniphysgs}, PhysSplat \citep{zhao2025physsplat}, PhysGM \citep{lv2026physgm}, NGFF \citep{li2026ngff}, and PhysGen3D \citep{chen2025physgen3d} integrate physics simulation with reconstructed 3D representations, while PhysDreamer \citep{zhang2024physdreamer}, PhysFlow \citep{liu2025physflow}, and Phys4D \citep{lu2026phys4d} use video generation models as motion priors or generative backbones to synthesize plausible 4D dynamics. These works validate physics priors for reconstruction, but generative models \citep{yang2025cogvideox,zhang2025tora} remain difficult to control through physical interaction inputs, and physical digital twins can be limited by fixed substrates, pure parameter fitting, or brittle action grounding. Closest to our setting, PhysTwin and its extensions \citep{jiang2025phystwin,zhang2025real2simeval} and Spring-Gaus \citep{springgaus2024} couple physics substrates with Gaussian splatting from videos, and EMPM \citep{chen2026empm} fits differentiable MPM for manipulation. NCLaw \citep{ma2023nclaw}, MASIV \citep{zhao2025masiv}, and NeuMA \citep{cao2024neuma} learn or adapt neural constitutive models, and VisionLaw \citep{lin2026visionlaw} infers symbolic constitutive laws from visual observations. We instead use physics-neural dynamics as the core transition model, together with compliant distributed actuation and heterogeneous constitutive modeling, to turn reconstruction into interactive modeling.

\paragraph{Neural dynamics of deformable objects.}
Learning-based simulators replace analytical dynamics with neural transition models \citep{ai2025dynamicsreview}. Particle graph \citep{sanchez2020gns} and particle-grid networks \citep{zhang2025pgnd} model ropes, cloths, and volumetric objects; GS-Dynamics \citep{zhang2024gsdynamics} couples Gaussian tracking with graph dynamics, and AdaptiGraph \citep{zhang2024adaptigraph} conditions graph dynamics on physical-property estimates. These methods are flexible, but purely learned transitions often need substantial data, remain tied to the training distribution, and drift under long rollouts or novel actions. We instead use a physics-guided rollout for stronger generalization, with neural dynamics acting as a residual correction for real-world mismatch.

\paragraph{Hybrid physics-generative world models.}
Recent systems also use physics to support action-conditioned 4D world prediction. Building on the static-scene precursor WonderWorld \citep{yu2024wonderworld}, the Wonder series couples physics solvers with video generation for interactive content creation through WonderPlay \citep{li2025wonderplay}, PerpetualWonder \citep{zhan2026perpetualwonder}, and RealWonder \citep{liu2026realwonder}. Force Prompting \citep{gillman2025forceprompt} and Goal Force \citep{gillman2026goalforce} further fine-tune video diffusion models on synthetic physics primitives to absorb force control signals. These methods show that physics can guide video generation, but require large-scale generative-model training. We instead learn from real videos the underlying physics in a data-efficient way.

\section{DeformMaster}
\label{sec:method}

\begin{figure}[t]
    \centering
    \includegraphics[width=\textwidth]{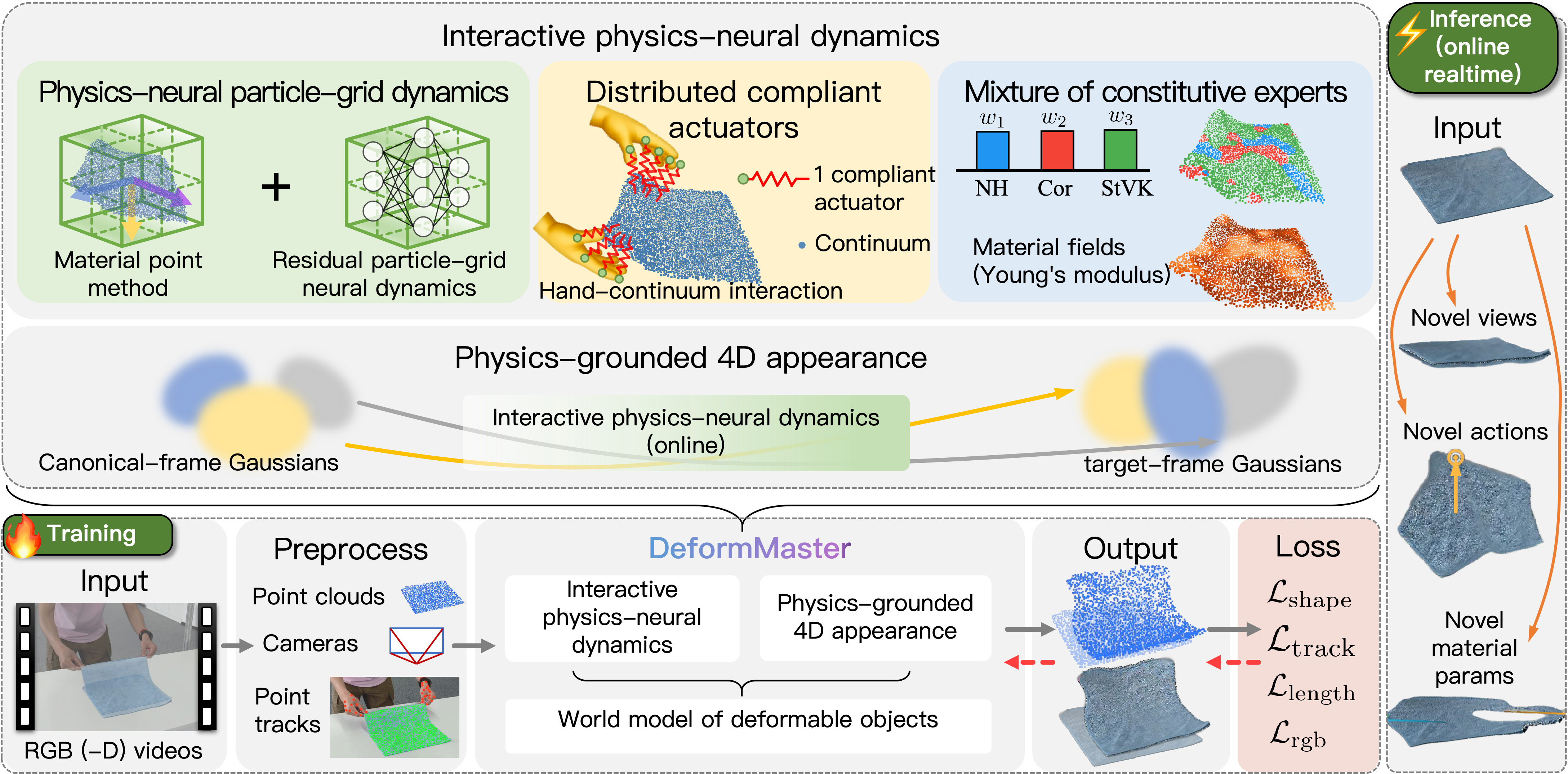}
    \caption{Overview of DeformMaster. From interaction videos, DeformMaster unifies interactive physics-neural dynamics and physics-grounded 4D appearance in a single deformable-object world model. The dynamics module integrates Physics-Neural Particle-Grid Dynamics, Distributed Compliant Actuators, and a Mixture of Constitutive Experts to support stable rollout, robust hand-continuum interaction, and heterogeneous material response. Within the dynamics module, actuators provide accelerations, and constitutive experts provide stresses for the material point method update. Gaussian splats are driven by the dynamics branch. Training optimizes dynamics and appearance using geometric, tracking, and image supervision. Inference supports novel view, action, and material-parameter conditions online.}
    \label{fig:method}
\end{figure}

We seek an instance-specific world model of deformable objects that (i) rolls out stable dynamics aligned with real-world observations, (ii) grounds noisy hand tracks for effective hand-continuum interaction, (iii) captures complex material response, and (iv) renders high-fidelity appearance grounded in physics. The model is optimized separately for each observed interaction sequence.
To this end, as illustrated in \cref{fig:method}, DeformMaster pairs interactive physics-neural dynamics (\cref{sec:method:dynamics}) with physics-grounded appearance (\cref{sec:method:appearance}) through four components: PNPGD for dynamics rollout, DCA for interaction, MoCE for material response, and Gaussian Splatting for rendering. Physics preliminaries are provided in Appendix~\ref{app:preliminaries}.


\subsection{Problem Formulation}
\label{sec:method:problem}

\textbf{State.} We represent the deformable state as $s_t=(s^{\mathrm{mat}}_t,s^{\mathrm{app}}_t)$, where $s^{\mathrm{mat}}_t$ denotes the material-particle state and $s^{\mathrm{app}}_t$ the appearance-particle state. 
\textbf{Actions.} The action $a_t$ consists of observed hand or actuator anchor positions and velocities.
\textbf{Observations.} Supervision comes from monocular RGB or multi-view RGB-D videos with extracted point clouds, camera poses, and dense 3D tracks. 
\textbf{Learning objective.} We learn $(\theta,\phi,\psi)$ for the joint dynamics-appearance model:
\begin{equation}
    s^{\mathrm{mat}}_{t+1} \;=\; \mathcal{F}_{\theta,\phi}(s^{\mathrm{mat}}_t,\, a_t),
    \qquad
    \hat{I}_{t+1} \;=\; \mathcal{G}_\psi(s^{\mathrm{app}}_{t+1}) \;=\; \mathcal{G}_\psi\!\big(\mathcal{B}_\psi(s^{\mathrm{mat}}_{t+1})\big),
    \label{eq:problem:transition}
\end{equation}
where $\mathcal{F}_{\theta,\phi}$ is the hybrid physics-neural rollout operator for material dynamics; $\mathcal{B}_\psi$ bridges material particles to appearance particles; and $\mathcal{G}_\psi$ renders that state into images. Training matches both rolled-out material states and rendered frames to observations.

\subsection{Interactive Physics-Neural Dynamics}
\label{sec:method:dynamics}

\paragraph{Physics-Neural Particle-Grid Dynamics (PNPGD).}
\label{sec:method:pnpgd}
Physics-based simulators provide structured, stable priors for dynamics, but real observations exhibit systematic effects that no idealized model can fully express. We therefore pair an explicit physics block with a neural residual that absorbs the unmodeled mismatch.

We decompose deformation dynamics $\mathcal{F}_{\theta,\phi}$ into a physics block $\mathcal{P}_\theta$ and a residual block $\mathcal{R}_\phi$:
\begin{equation}
    \mathcal{F}_{\theta,\phi}
    \;=\;
    \mathcal{P}_\theta \oplus \mathcal{R}_\phi.
    \label{eq:hybrid_general}
\end{equation}
Concretely, $\mathcal{P}_\theta$ first advances the state over one frame with differentiable MPM~\citep{hu2018mlsmpm} under material parameters $\theta$, producing a tentative next state (with $s_t$ denoting $s^{\mathrm{mat}}_t$ for brevity):
\begin{equation}
    \tilde{s}_{t+1} \;=\; \mathcal{P}_\theta^{\text{MPM}}\!\big(s_t,\, a_t\big),
    \label{eq:hybrid_phys}
\end{equation}
where $s_t = (\mathbf{x}_p^t, \mathbf{v}_p^t, \mathbf{F}_p^t, \mathbf{C}_p^t)$ is the MPM particle state consisting of per-particle position, velocity, deformation gradient, and affine matrix; $a_t$ denotes the action (detailed in DCA below). Second, after the MPM rollout over one frame, a residual block $\mathcal{R}_\phi$ predicts a neural velocity correction $\Delta\mathbf{v}_p$,
\begin{equation}
    \Delta\mathbf{v}_p \;=\; \mathcal{R}_\phi\!\big(\tilde{s}_{t+1},\, s_t,\, h_t\big),
    \label{eq:hybrid_residual}
\end{equation}
where $h_t = \{(\mathbf{x}_p^{t-i}, \mathbf{v}_p^{t-i})\}_{i=1}^{H}$ is a short kinematic history. The final residual-corrected state $s_{t+1}$ is then given by
\begin{equation}
    \mathbf{v}_p^{t+1} \;=\; \tilde{\mathbf{v}}_p^{t+1} + \Delta\mathbf{v}_p,
    \qquad
    \mathbf{x}_p^{t+1} \;=\; \tilde{\mathbf{x}}_p^{t+1} + \Delta\mathbf{v}_p \, \Delta t,
    \label{eq:hybrid_update}
\end{equation}
where the residual updates only particle positions and velocities, while $\mathbf{F}_p^{t+1}$ and $\mathbf{C}_p^{t+1}$ are inherited from $\tilde{s}_{t+1}$. By design $\|\Delta\mathbf{v}_p\|$ is bounded, so $\mathcal{R}_\phi$ acts as a perturbation at the frame level rather than a free state predictor.

To compose cleanly with MPM, residual block $\mathcal{R}_\phi$ uses a similar particle-grid representation. We reformulate Particle-Grid Neural Dynamics \citep{zhang2025pgnd} as this residual block: a PointNet encoder \citep{qi2017pointnet} produces a per-particle latent feature
\begin{equation}
    \mathbf{f}_p \;=\; E_\phi\!\big(\tilde{\mathbf{x}}_p^{t+1},\, \tilde{\mathbf{v}}_p^{t+1},\, \tilde{\mathbf{x}}_p^{t+1} - \mathbf{x}_p^{t},\, h_t\big),
    \label{eq:residual_pgnd_encoder}
\end{equation}
that summarises the post-MPM state and history $h_t$; a coordinate-conditioned MLP decoder with Fourier positional encoding \citep{mildenhall2020nerf} predicts a bounded grid-node correction $\mathbf{u}_g = \alpha\,\tanh(\cdot)$, which is then mapped back to particles by the same B-spline weights MPM uses for P2G/G2P transfers, yielding $\Delta\mathbf{v}_p$. The residual architecture mirrors MPM's particle-grid hybrid through an Eulerian grid representation and particle-grid transfer. Details of the MPM configuration and neural residual architecture are
provided in \cref{app:mpm_config,app:residual_pgnd_arch}.


\paragraph{Distributed Compliant Actuator (DCA).}
\label{sec:method:dca}

Hand-continuum interaction is often the fragile part of real-to-sim deformable modeling. Vision-derived hand or actuator tracks are noisy and sparse, so hard pointwise constraints surface two failure modes: tracking noise is injected as velocity spikes, and point loads deform only a tiny neighborhood. 
DCA addresses these failure modes with \emph{compliance}, which turns hard constraints into compliant actuator-particle couplings absorbing high-frequency noise, and \emph{distribution}, which spreads actuation over a finite contact patch to drive bulk motion in a soft continuum.

Concretely, DCA applies compliant actuator-to-particle couplings over a local actuator neighborhood, producing the actuator-induced acceleration $\mathbf{a}_p$ on particle $p$:
\begin{equation}
    \mathbf{a}_p
    \;=\;
    n_p^{-1/2}
    \sum\nolimits_{c \in \mathcal{N}(p)}
    \Big[
        k_p\big((\mathbf{x}_c - \mathbf{x}_p) - \mathbf{o}_{p,c}^0\big)
        + k_d\big(\mathbf{v}_c - \mathbf{v}_p\big)
    \Big],
    \label{eq:dca}
\end{equation}
where $k_p,k_d$ are stiffness and damping (DCA gains), $\mathbf{x}_p,\mathbf{v}_p$ are the particle state, $\mathbf{x}_c,\mathbf{v}_c$ are actuator-anchor state, $\mathbf{o}_{p,c}^0$ is the initial rest offset, $\mathcal{N}(p)$ is the local actuator neighborhood, $n_p=|\mathcal{N}(p)|$. The $n_p^{-1/2}$ factor normalizes over multiple anchors to avoid force stacking.


\paragraph{Mixture of Constitutive Experts (MoCE).}
\label{sec:method:moe}

Real deformable objects rarely conform to a single idealized constitutive law. Their response is shaped by material composition, processing history, and scene-specific deformation patterns. We therefore represent stress with a finite mixture of constitutive experts whose weights can adapt across continuum regions.
We model the first Piola--Kirchhoff stress as a spatially varying mixture of constitutive experts:
\begin{equation}
    \mathbf{P}_{\text{mix}}(\mathbf{F}_p; E_p, \nu_p)
    \;=\;
    \sum\nolimits_k w_{k,p} \, \mathbf{P}_k(\mathbf{F}_p; E_p, \nu_p),
    \label{eq:moe}
\end{equation}
where $\mathbf{P}_k$ is the stress map of expert $k\in\{\text{NH, Cor, StVK}\}$, $\mathbf{F}_p$ is the deformation gradient of particle $p$, and $E_p,\nu_p$ denote learnable Young's modulus and Poisson's ratio. The mixture weights $w_{k,p}$ are spatially varying, implemented through patch-level expert logits and interpolated to particles as detailed in \cref{app:moe_patch_param}.


\subsection{Physics-Grounded 4D Appearance}
\label{sec:method:appearance}

Online interaction requires high-fidelity rendering without re-optimizing a dynamic appearance model after every new action. We therefore keep rollout on a compact material-particle state and use it to drive Gaussian appearance. Given the particle trajectory predicted by $\mathcal{F}_{\theta,\phi}$, the bridge $\mathcal{B}_\psi$ deforms Gaussians using LBS \citep{sumner2007embedded,huang2024scgs}. The update is incremental: each frame applies the particle motion from the previous frame to the current one, rather than re-skinning from the canonical state. This keeps rendering efficient and aligned with physical motion without learning a separate dynamic reconstruction model.

\subsection{Training Scheme}
\label{sec:method:training}

We train DeformMaster with a combined dynamics-and-appearance objective:
\begin{equation}
    \mathcal{L}_\text{total}
    =
    \mathcal{L}_\text{dynamics}
    +
    \mathcal{L}_\text{appearance}
    =
    \lambda_\text{track}\,\mathcal{L}_\text{track}
    + \lambda_\text{shape}\,\mathcal{L}_\text{shape}
    + \lambda_\text{len}\,\mathcal{L}_\text{len}
    + \lambda_\text{rgb}\,\mathcal{L}_\text{rgb},
\end{equation}
where $\mathcal{L}_\text{dynamics}$ contains track, shape, and length terms: $\mathcal{L}_\text{track}$ aligns 3D point trajectories, $\mathcal{L}_\text{shape}$ aligns shape with Chamfer distance, and $\mathcal{L}_\text{len}$ preserves local lengths. $\mathcal{L}_\text{appearance}$ is $\mathcal{L}_\text{rgb}$, a photometric loss on rendered frames.
Optimization proceeds in multiple stages. Stages~1--2 alternate between dynamics training and DCA gain selection: we first learn the material fields and neural residual with default DCA gains under dynamics loss, use the warm-started dynamics to select gains with CMA-ES~\citep{hansen2006cma}, and then continue learning the material fields and neural residual with the updated gains. Stage~3 first optimizes Gaussian splats with appearance loss and then uses the total loss (\textit{i.e.}, adding RGB loss) to refine the dynamics branch.

\section{Experiments}
\label{sec:experiments}

\subsection{Setup}
\label{sec:experiments:setup}

We organize the evaluation around the contributions of DeformMaster. \textbf{Implementation.} Implementation details (preprocessing, MPM, neural residual, MoCE, and training) are deferred to \cref{sec:appendix}. \textbf{Dataset.} We evaluate on 20 real deformable-object sequences from PhysTwin \citep{jiang2025phystwin}, spanning deformable linear ($n{=}3$, ropes), planar ($n{=}9$, cloths and package), and volumetric ($n{=}8$, soft-bodied toys) objects, all captured with calibrated three-view RGB-D videos at 30\,fps. For future prediction, we train on the first 70\% of each sequence and evaluate on the remaining 30\%. \textbf{Baselines.} We compare the full system against PhysTwin \citep{jiang2025phystwin}, Spring-Gaus \citep{springgaus2024}, and GS-Dynamics \citep{zhang2024gsdynamics}, and ablate PNPGD, DCA, MoCE, and RGB-guided dynamics refinement (\cref{sec:experiments:main,sec:ablations}). \textbf{Metrics.} We evaluate future rollout with dynamics metrics (Chamfer distance, Track error, and mask IoU) and appearance metrics (PSNR, SSIM, and LPIPS). 
\textbf{Online playground.} Our method supports online dynamics rollout at over $15$\,fps; the online playground is shown in \cref{app:online_playground}, with additional interactive results provided on the project page. Code for the playground is publicly available at {\urlstyle{same}\url{https://can-lee.github.io/deformmaster-web/}}.

\subsection{Main Results}
\label{sec:experiments:main}

\begin{table}[t]
    \centering
    \caption{Overall future-prediction comparison on twenty real-world PhysTwin deformation sequences. We report dynamics accuracy and rendered appearance quality averaged over unseen test frames. DeformMaster achieves the strongest overall performance, improving dynamics and appearance fidelity while remaining comparable to PhysTwin on Track error.}
    \label{tab:main_summary}
    \small
    \setlength{\tabcolsep}{4pt}
    \begin{tabular}{l ccc ccc}
        \toprule
        \multirow{2}{*}{Method} & \multicolumn{3}{c}{Future dynamics} & \multicolumn{3}{c}{Future appearance} \\
        \cmidrule(lr){2-4}\cmidrule(lr){5-7}
        & IoU $\uparrow$ & Chamfer $\downarrow$ & Track $\downarrow$ & PSNR $\uparrow$ & SSIM $\uparrow$ & LPIPS $\downarrow$ \\
        \midrule
        PhysTwin \citep{jiang2025phystwin}       & 0.734 & 0.012 & \textbf{0.023} & 25.16 & 0.935 & 0.061 \\
        Spring-Gaus \citep{springgaus2024}       & 0.464 & 0.062 & 0.094 & 22.49 & 0.924 & 0.113 \\
        GS-Dynamics \citep{zhang2024gsdynamics}  & 0.498 & 0.041 & 0.070 & 22.54 & 0.924 & 0.097 \\
        \midrule
        \textbf{DeformMaster (ours)}             & \textbf{0.748} & \textbf{0.011} & 0.024 & \textbf{25.41} & \textbf{0.936} & \textbf{0.061} \\
        \bottomrule
    \end{tabular}
\end{table}

\begin{table}[t]
    \centering
    \caption{Per-category dynamics prediction on the PhysTwin sequences, grouped into deformable linear, planar, and volumetric objects. DeformMaster improves on linear and volumetric objects, while the remaining Track gap is concentrated in planar object sequences.}
    \label{tab:Per-category_dynamics}
    \small
    \setlength{\tabcolsep}{3.2pt}
    \resizebox{\linewidth}{!}{%
    \begin{tabular}{l ccc ccc ccc}
        \toprule
        \multirow{2}{*}{Method} & \multicolumn{3}{c}{Linear ($n{=}3$)} & \multicolumn{3}{c}{Planar ($n{=}9$)} & \multicolumn{3}{c}{Volumetric ($n{=}8$)} \\
        \cmidrule(lr){2-4}\cmidrule(lr){5-7}\cmidrule(lr){8-10}
        & IoU $\uparrow$ & Chamfer $\downarrow$ & Track $\downarrow$ & IoU $\uparrow$ & Chamfer $\downarrow$ & Track $\downarrow$ & IoU $\uparrow$ & Chamfer $\downarrow$ & Track $\downarrow$ \\
        \midrule
        PhysTwin \citep{jiang2025phystwin}       & 0.658 & 0.007 & 0.013 & 0.738 & 0.013 & \textbf{0.028} & 0.748 & 0.013 & 0.021 \\
        \textbf{DeformMaster (ours)}             & \textbf{0.721} & \textbf{0.005} & \textbf{0.010} & \textbf{0.748} & \textbf{0.013} & 0.032 & \textbf{0.756} & \textbf{0.012} & \textbf{0.020} \\
        \bottomrule
    \end{tabular}%
    }
\end{table}

\Cref{tab:main_summary} reports future-frame prediction results on the 20 PhysTwin sequences. Our DeformMaster achieves the strongest overall performance, improving mask IoU and Chamfer distance over all baselines and producing the best rendered appearance. Compared with PhysTwin, our method slightly improves Chamfer ($0.011$ vs.\ $0.012$) and IoU ($0.748$ vs.\ $0.734$), while maintaining a comparable Track error ($0.024$ vs.\ $0.023$). The advantage over learning-based Gaussian dynamics baselines is more pronounced: our method reduces Chamfer by more than $3\times$ and Track error by at least $2.9\times$ compared with Spring-Gaus and GS-Dynamics. Since the appearance metrics are computed by deforming the same first-frame Gaussian representation with the predicted trajectory, the improvements in PSNR, SSIM, and LPIPS mainly reflect better long-horizon rollout rather than renderer-specific tuning.

\Cref{tab:Per-category_dynamics} further breaks down the dynamics results by object type. The gains are most pronounced on linear objects (ropes), where our DeformMaster improves IoU from $0.658$ to $0.721$ and reduces Chamfer and Track error from $0.007/0.013$ to $0.005/0.010$. On volumetric objects (soft-bodied toys), our method also improves all three dynamics metrics, indicating that the particle-grid-based transition model benefits objects with substantial three-dimensional deformation. Planar objects are the most challenging case: our method slightly improves IoU and matches Chamfer, but its Track error is higher than PhysTwin ($0.032$ vs.\ $0.028$). This suggests that the small overall Track gap in \cref{tab:main_summary} is mainly driven by planar sequences, where single-layer cloth-like motion is less naturally matched to the particle-grid continuum simulator.

\begin{figure}[t]
    \centering
    \includegraphics[width=\linewidth]{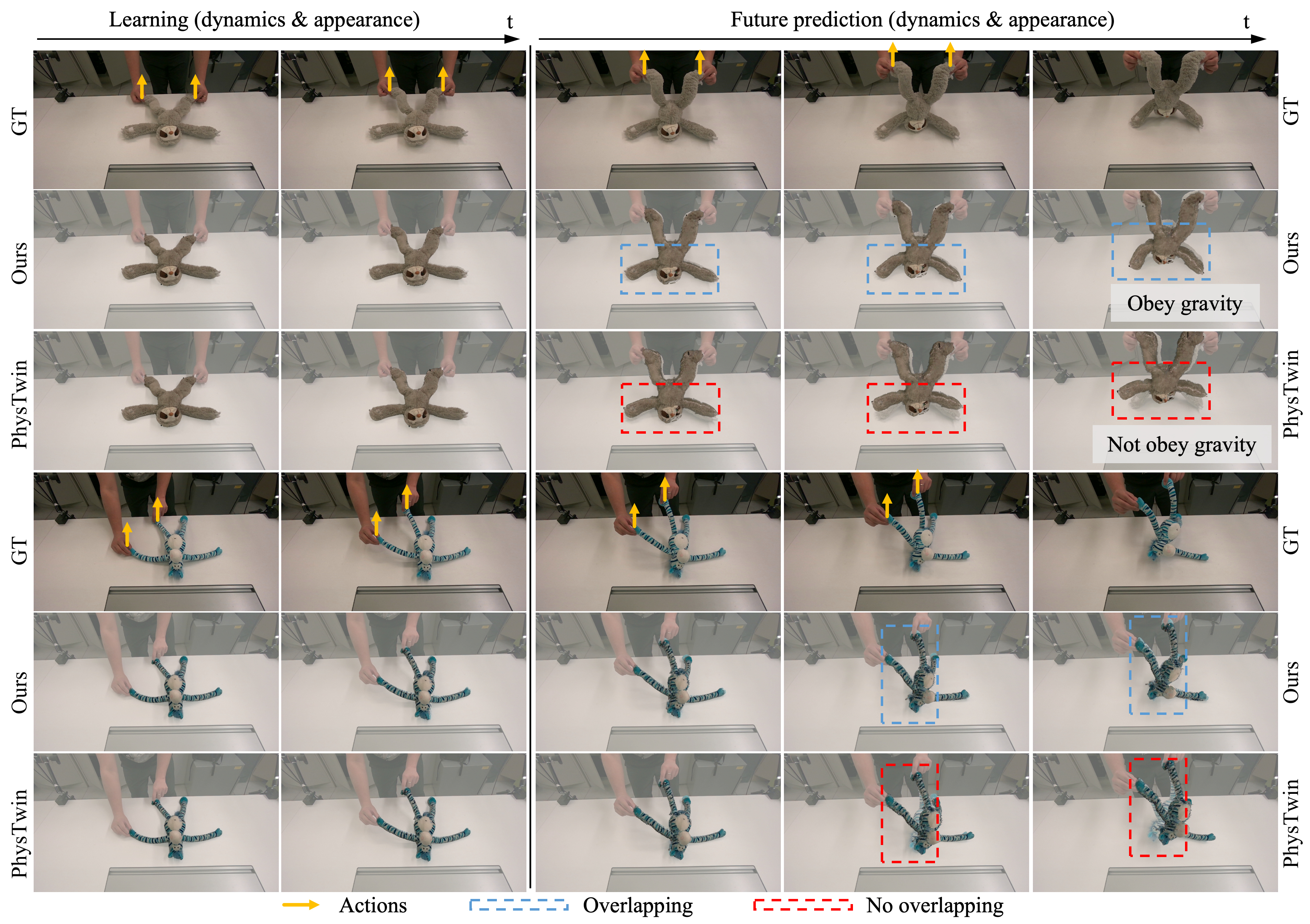}
    \caption{Qualitative comparison on two representative deformation sequences. Columns progress from the learning stage to future prediction, and each case compares ground truth (GT), DeformMaster (Ours), and PhysTwin. Yellow arrows indicate the applied actions. Blue and red dashed boxes mark corresponding regions in our prediction and PhysTwin, respectively. DeformMaster better preserves the future object configuration and yields stronger overlap with GT in the highlighted regions.}
    \label{fig:qualitative_phystwin}
\end{figure}

\Cref{fig:qualitative_phystwin} visualizes two representative deformable volumetric objects from the PhysTwin sequences. During the learning stage, both methods capture the dynamics and appearance well. In future prediction, our method better preserves the object configuration under upward pulling and gravity, and remains more closely aligned with the ground truth in the highlighted regions. These results qualitatively echo the design goals: stable long-horizon rollouts, compensation for idealized-physics mismatch, effective actuation from noisy contacts, and appearance tied to the predicted physical state. In contrast, PhysTwin tends to under-deform or drift away from the observed shape, leading to weaker overlap in the same regions.

\begin{table}[t]
    \centering
    \caption{Quantitative dynamics and appearance prediction under novel actions. We use object dynamics learned from one interaction sequence to predict responses to unseen action trajectories from another sequence of the same object, without retraining or adaptation.}
    \label{tab:novel_action}
    \small
    \setlength{\tabcolsep}{4pt}
    \begin{tabular}{l ccc ccc}
        \toprule
        Method & IoU $\uparrow$ & Chamfer $\downarrow$ & Track $\downarrow$ & PSNR $\uparrow$ & SSIM $\uparrow$ & LPIPS $\downarrow$ \\
        \midrule
        PhysTwin \citep{jiang2025phystwin} & 0.5837 & 0.01954 & 0.03355 & 23.40 & 0.9308 & 0.0741 \\
        \textbf{DeformMaster (ours)} & \textbf{0.6326} & \textbf{0.01882} & \textbf{0.03299} & \textbf{23.62} & \textbf{0.9311} & \textbf{0.0721} \\
        \bottomrule
    \end{tabular}
\end{table}

To evaluate generalization to unseen actions, we use object dynamics learned from one interaction sequence to predict responses to new action trajectories from another sequence of the same object. For both methods, all learned parameters remain fixed, and the target sequence provides only the initial geometry and controller trajectory, with no target-action training or adaptation. Predictions are evaluated against target observations on the future-test split. \Cref{tab:novel_action} shows that DeformMaster improves all six dynamics and appearance metrics over PhysTwin, demonstrating its ability to predict responses to unseen actions without retraining. Evaluation details are provided in \cref{app:novel_action_sequences}. We also evaluate view generalization by rendering predicted future dynamics from unseen viewpoints. As shown in \cref{tab:novel_view}, DeformMaster achieves higher PSNR and SSIM and lower LPIPS than PhysTwin.

\begin{figure}[!t]
    \centering
    \includegraphics[width=\linewidth]{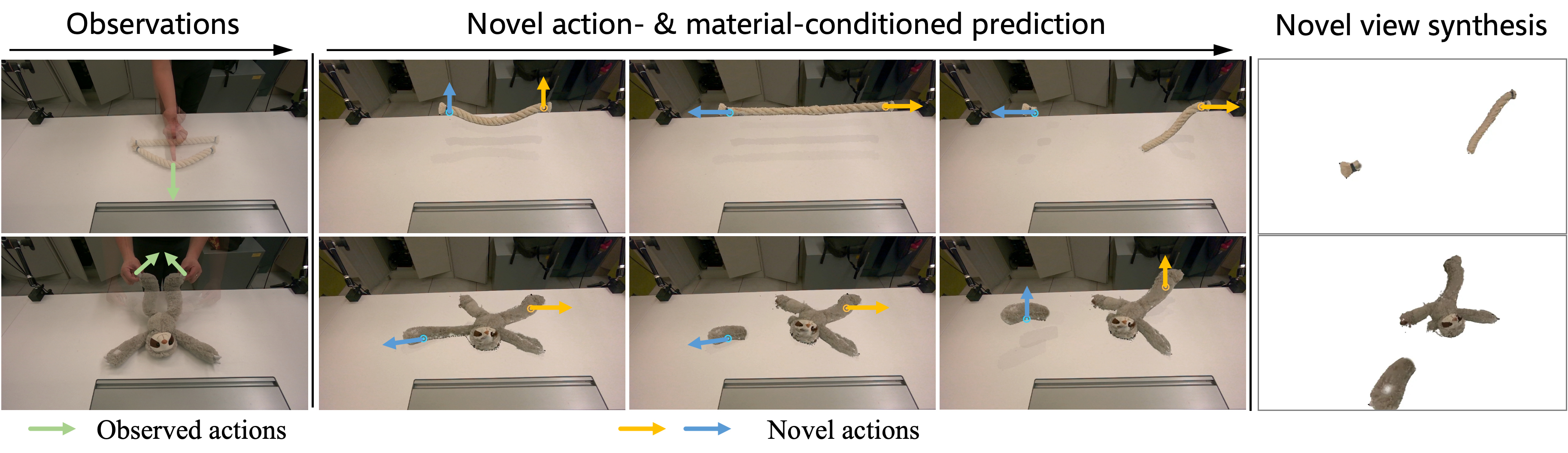}
    \caption{Generalization to novel action-, material-, and view-conditioned prediction with our DeformMaster. DeformMaster supports rollouts under novel actions and material scales. Scaling the material fields to $0.3\times$ their learned values produces fracture behavior, while the final column renders the fracture states from a novel view, demonstrating interactive ability beyond the observations and baselines.}
    \label{fig:qualitative_novelcond}
\end{figure}

\begin{figure}[!t]
    \centering
    \includegraphics[width=\linewidth]{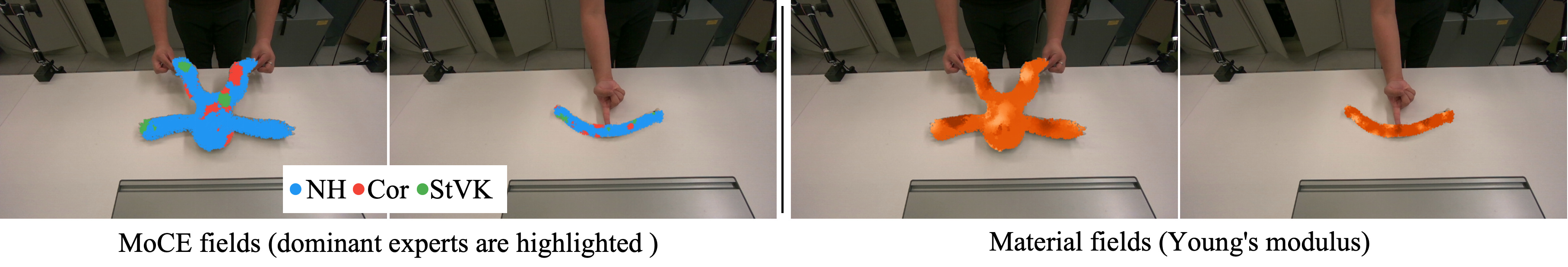}
    \caption{Qualitative visualization of MoCE and material fields. For readability, the left two panels visualize the MoCE mixture by showing only the largest-weight constitutive expert at each particle. The right two panels visualize the material field as the spatial distribution of Young's modulus. The results qualitatively show that our method captures spatially varying material behavior for deformable objects, rather than assuming a single homogeneous constitutive response.}
    \label{fig:materials}
\end{figure}

\Cref{fig:qualitative_novelcond} shows that our model can be queried under novel action, material parameters, and view conditions. Starting from the same observed object, our method performs rollouts under novel conditions by changing the actuation direction and material parameter scale, and then renders the predicted state from a novel camera view. Notably, scaling the material fields to $0.3\times$ their recovered values produces material fracture, illustrating a discontinuous behavior that is difficult for PhysTwin's fixed topological connectivity to express.

\Cref{fig:materials} visualizes MoCE and material fields on representative deformable objects. In MoCE, each particle uses a mixture over constitutive experts; for readability, the visualization shows only the dominant expert with the largest mixture weight at each particle. The displayed expert maps and the corresponding Young's modulus fields are spatially non-uniform, indicating that the optimized material response adapts across object regions. Together, these visualizations show that our model captures region-dependent material behavior from video observations, rather than reducing each deformable object to a single homogeneous constitutive response.

\subsection{Ablation Study}
\label{sec:ablations}

\begin{figure}[t]
    \centering
    \includegraphics[width=\textwidth]{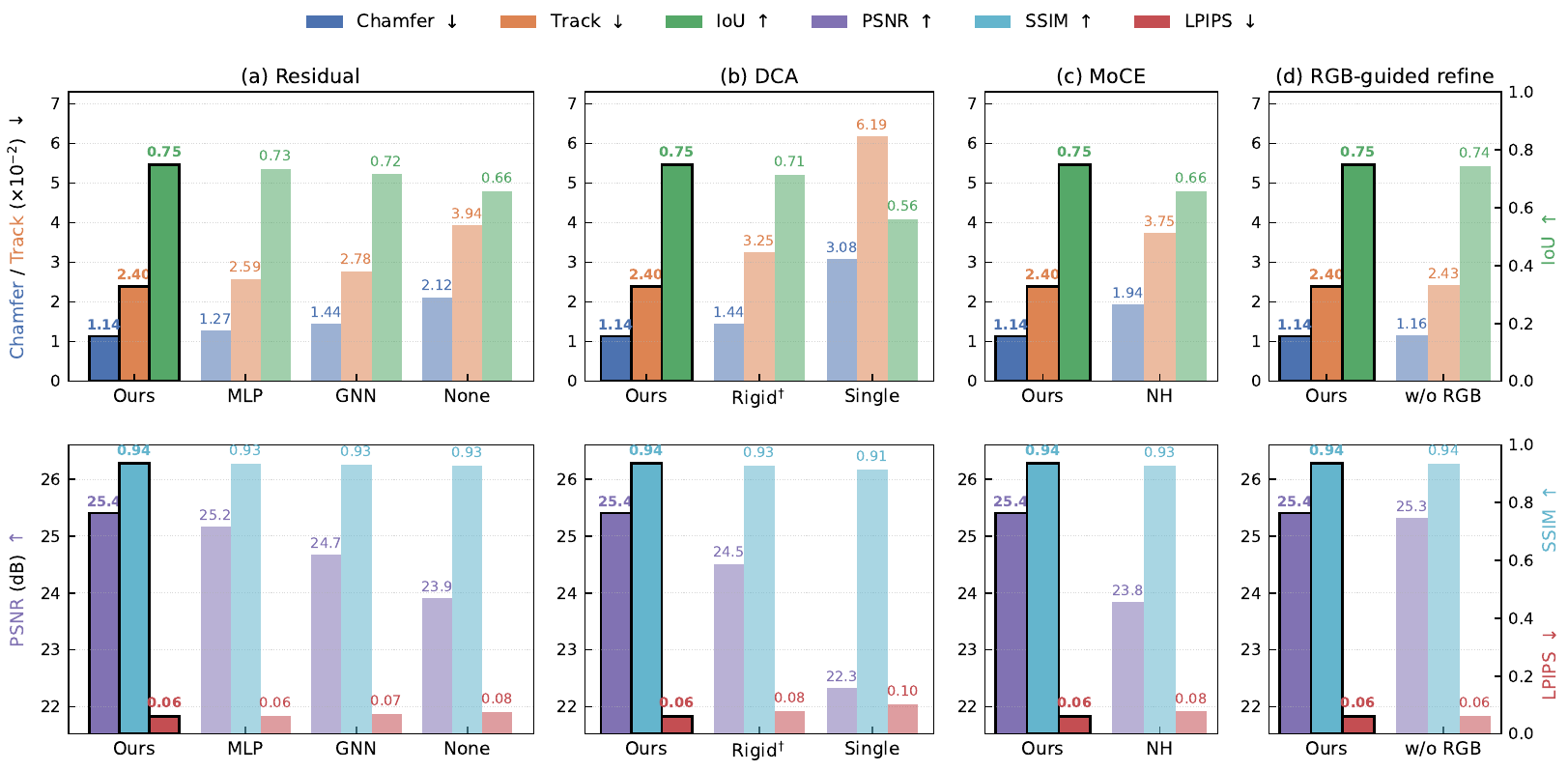}
    \caption{Ablation summary. Columns (a)--(d) compare four design choices, with dynamics metrics in the top row (Chamfer/Track in $10^{-2}$, IoU) and appearance metrics in the bottom row (PSNR, SSIM, LPIPS). \textbf{(a) Residual}: the neural particle-grid residual performs best, while MLP/GNN residuals and removing the residual (None, \textit{i.e.}, MPM only) degrade both dynamics rollout and appearance. \textbf{(b) DCA}: removing compliance (Rigid$^{\dagger}$, reported on $11/20$ successfully trained sequences) or collapsing the distributed actuator to a single one (Single) both reduce performance, showing the need for distributed compliant actuation. \textbf{(c) MoCE}: replacing the mixture with a single Neo-Hookean expert worsens both dynamics and appearance. \textbf{(d) RGB-guided refinement}: fine-tuning with RGB loss in training stage~3 improves dynamics metrics and rendered appearance over the dynamics-loss-only variant (w/o RGB). Overall, each component contributes to the final system; ``Ours'' is highlighted with a solid edge, and full numerical results are in \cref{app:ablation_tables}.}
    \label{fig:ablation_summary}
\end{figure}

\paragraph{PNPGD: Residual Matters, and Its Architecture Matters.}
\label{sec:abl_res}

\Cref{fig:ablation_summary}(a) ablates PNPGD by varying the residual branch. We compare the proposed particle-grid residual with three alternatives: removing the residual entirely, using an MLP residual that predicts independent per-particle corrections, and using a GNN residual in the style of particle-based neural simulators~\citep{sanchez2020gns}. The MPM-only variant without residual degrades both dynamics and appearance, confirming that the pure physics simulator provides a useful prior but cannot by itself absorb the systematic mismatch present in real videos. Among residual variants, both the MLP and GNN underperform the proposed design, showing that the particle-grid residual architecture is more effective. Together, these results support the motivation in \cref{sec:method:pnpgd}: a neural residual is needed to bridge the gap between idealized physics and real-world observations, and the particle-grid residual architecture matters. Full numbers are reported in \cref{tab:ablation_residual_overall,tab:ablation_residual}.

\paragraph{DCA: Both Compliance and Distribution Matter.}
\label{sec:abl_dca}

\Cref{fig:ablation_summary}(b) ablates the two design choices in DCA. The rigid variant removes compliance by replacing compliant actuator-particle coupling with hard constraints, while the single-actuator variant keeps compliance but collapses each distributed actuator to one actuator. The rigid variant diverges on $9/20$ sequences, including all $8$ volumetric cases, which is consistent with the motivation that noisy vision-derived tracks should not be injected as hard pointwise constraints. The single-actuator variant is numerically stable but reduces accuracy, showing that pointwise forcing is insufficient to drive bulk deformation in a soft continuum. These results verify the two-part DCA design in \cref{sec:method:dca}: compliance stabilizes action grounding under noisy tracks, and distribution improves force transmission over contact regions of deformable objects. Full numbers are reported in \cref{tab:ablation_dca_overall,tab:ablation_dca}.

\paragraph{MoCE vs.~Single Expert.}
\label{sec:abl_moe}

\Cref{fig:ablation_summary}(c) ablates the constitutive model by replacing MoCE with a single Neo-Hookean expert while keeping the rest of the pipeline unchanged. This removes the spatially varying constitutive experts and forces all particles to share the same canonical constitutive law. The single-expert variant worsens both rollout accuracy and rendered appearance, indicating that real deformation sequences contain heterogeneous material responses that cannot be fully represented by a global constitutive choice. In contrast, MoCE represents the stress response as a spatially varying mixture of canonical constitutive experts, allowing the model to adapt across continuum regions while remaining grounded in analytic constitutive laws. Full numbers are reported in \cref{tab:ablation_moe_overall,tab:ablation_moe}.

\paragraph{Effect of RGB-Guided Refinement.}
\label{sec:abl_rgb_refine}

\Cref{fig:ablation_summary}(d) evaluates RGB-guided dynamics refinement by ablating the RGB loss in training stage~3. We compare whether to use the total loss (\textit{i.e.}, adding RGB loss) to refine the dynamics branch. Because the rendered appearance is driven by the predicted material-particle trajectory (\cref{eq:problem:transition}), the RGB loss provides supervision to the underlying dynamics rather than only to image-space appearance. Adding this loss yields consistent overall gains, improving both dynamics and rendered appearance metrics. Per-category  results in \cref{tab:stage3_ablation} show that the dynamics gains mainly come from deformable planar and volumetric objects, while linear objects remain nearly unchanged. These results show that RGB supervision in stage~3 contributes to rollout refinement. Full numbers are reported in \cref{tab:stage3_ablation_overall,tab:stage3_ablation}.

\subsection{Runtime, Memory, and Scalability}
\label{sec:runtime}

To assess the computational cost of online interaction, we profile DeformMaster on a single NVIDIA A100 GPU. Dynamics rollout for the particle-heavy ``double\_\allowbreak{}lift\_\allowbreak{}sloth'' case takes 62.674\,ms/frame (15.99\,FPS; \cref{tab:runtime_breakdown}). Including appearance computation brings the estimated total latency to 75.612\,ms/frame (13.23\,FPS). MPM and the neural residual account for 82.5\% of the total latency, versus 2.18\% for Gaussian rendering. Separately measured peak memory is 2.263\,GB for dynamics and 1.859\,GB for appearance.

\Cref{tab:runtime_scaling} reports dynamics-only scaling. Increasing substeps from 21 to 84 raises median latency from 42.184 to 111.333\,ms/frame; total rollout time increases with the horizon. Increasing particles from 9,133 to 22,132 raises latency and memory, while grid resolution has little effect over the tested $32$--$64$ range. Peak dynamics memory stays within approximately 2.3\,GB.

Preprocessing takes minutes, while training requires approximately 2.5 hours of dynamics learning, DCA gain selection, and approximately 2--3 hours of appearance learning and RGB-guided refinement, with peak training memory around 20\,GB (\cref{tab:offline_costs}). These one-time per-instance costs are separate from online simulation and rendering.

\section{Conclusion}
\label{sec:conclusion}

We proposed DeformMaster, a video-derived interactive physics-neural world model for deformable objects. From real interaction videos, DeformMaster couples interactive physics-neural dynamics with physics-grounded appearance, enabling controllable rollout and high-fidelity rendering. This design reflects the central contributions: robust physics-neural dynamics, stable grounding of real interactions, heterogeneous material modeling, and 4D appearance synthesis driven by the underlying physics. Experiments on multi-category real sequences show the strongest overall future dynamics and appearance, with clear gains over baselines. Ablations further confirm that each component contributes to rollout accuracy and appearance quality. Limitations are discussed in \cref{sec:experiments:limitations}. Promising directions include extending DeformMaster to modeling under partial observations \citep{li2025orderedkeypoints}, contact-rich scenarios and fluid dynamics, and integrating language-conditioned goal prediction for robotic manipulation \citep{wu2025sport}.

\renewcommand{\acksection}{\section*{Acknowledgments}}
\begin{ack}
This work was supported by the National Natural Science Foundation of China (No. 62173192 and No. 62373365). This work was completed during an internship at Rightly Robotics, A4x. We thank Ren Li and Jie Gu for insightful discussions, Zhoujian Li for assistance, and Hanxiao Jiang for his excellent work PhysTwin and generous guidance.
\end{ack}

\bibliography{references}

@inproceedings{zhang2025pgnd,
  title       = {Particle-Grid Neural Dynamics for Learning Deformable Object Models from {RGB-D} Videos},
  author      = {Zhang, Kaifeng and Li, Baoyu and Hauser, Kris and Li, Yunzhu},
  booktitle   = {Proceedings of Robotics: Science and Systems},
  year        = {2025},
  doi         = {10.15607/RSS.2025.XXI.036},
}

@inproceedings{qi2017pointnet,
  title       = {{PointNet}: Deep Learning on Point Sets for {3D} Classification and Segmentation},
  author      = {Qi, Charles R. and Su, Hao and Mo, Kaichun and Guibas, Leonidas J.},
  booktitle   = {Proceedings of the IEEE Conference on Computer Vision and Pattern Recognition (CVPR)},
  year        = {2017},
  doi         = {10.1109/CVPR.2017.16},
  pages       = {77--85},
}

@inproceedings{mildenhall2020nerf,
  title       = {{NeRF}: Representing Scenes as Neural Radiance Fields for View Synthesis},
  author      = {Mildenhall, Ben and Srinivasan, Pratul P. and Tancik, Matthew and Barron, Jonathan T. and Ramamoorthi, Ravi and Ng, Ren},
  booktitle   = {Computer Vision -- ECCV 2020},
  year        = {2020},
  pages       = {405--421},
  doi         = {10.1007/978-3-030-58452-8_24},
}

@inproceedings{jiang2025phystwin,
  title       = {{PhysTwin}: Physics-Informed Reconstruction and Simulation of Deformable Objects from Videos},
  author      = {Jiang, Hanxiao and Hsu, Hao-Yu and Zhang, Kaifeng and Yu, Hsin-Ni and Wang, Shenlong and Li, Yunzhu},
  booktitle   = {Proceedings of the IEEE/CVF International Conference on Computer Vision (ICCV)},
  year        = {2025},
  pages       = {7219--7230},
  doi         = {10.1109/ICCV51701.2025.00678},
}

@inproceedings{springgaus2024,
  title       = {Reconstruction and Simulation of Elastic Objects with Spring-Mass {3D} Gaussians},
  author      = {Zhong, Licheng and Yu, Hong-Xing and Wu, Jiajun and Li, Yunzhu},
  booktitle   = {Computer Vision -- ECCV 2024},
  pages       = {407--423},
  year        = {2024},
  doi         = {10.1007/978-3-031-72627-9_23},
}

@inproceedings{zhang2024gsdynamics,
  title       = {Dynamic {3D} Gaussian Tracking for Graph-Based Neural Dynamics Modeling},
  author      = {Zhang, Mingtong and Zhang, Kaifeng and Li, Yunzhu},
  booktitle   = {Proceedings of The 8th Conference on Robot Learning},
  year        = {2025},
  pages       = {1851--1862},
  volume      = {270},
  series      = {Proceedings of Machine Learning Research},
  publisher   = {PMLR},
}

@inproceedings{sanchez2020gns,
  title       = {Learning to Simulate Complex Physics with Graph Networks},
  author      = {Sanchez-Gonzalez, Alvaro and Godwin, Jonathan and Pfaff, Tobias and Ying, Rex and Leskovec, Jure and Battaglia, Peter},
  booktitle   = {Proceedings of the 37th International Conference on Machine Learning},
  year        = {2020},
  pages       = {8459--8468},
  volume      = {119},
  series      = {Proceedings of Machine Learning Research},
  publisher   = {PMLR},
}

@inproceedings{li2023pacnerf,
  title       = {{PAC-NeRF}: Physics Augmented Continuum Neural Radiance Fields for Geometry-Agnostic System Identification},
  author      = {Li, Xuan and Qiao, Yi-Ling and Chen, Peter Yichen and Jatavallabhula, Krishna Murthy and Lin, Ming and Jiang, Chenfanfu and Gan, Chuang},
  booktitle   = {International Conference on Learning Representations (ICLR)},
  year        = {2023},
}

@inproceedings{xie2024physgaussian,
  title       = {{PhysGaussian}: Physics-Integrated {3D} Gaussians for Generative Dynamics},
  author      = {Xie, Tianyi and Zong, Zeshun and Qiu, Yuxing and Li, Xuan and Feng, Yutao and Yang, Yin and Jiang, Chenfanfu},
  booktitle   = {Proceedings of the IEEE/CVF Conference on Computer Vision and Pattern Recognition (CVPR)},
  year        = {2024},
  pages       = {4389--4398},
  doi         = {10.1109/CVPR52733.2024.00420},
}

@inproceedings{zhang2024physdreamer,
  title       = {{PhysDreamer}: Physics-Based Interaction with {3D} Objects via Video Generation},
  author      = {Zhang, Tianyuan and Yu, Hong-Xing and Wu, Rundi and Feng, Brandon Y. and Zheng, Changxi and Snavely, Noah and Wu, Jiajun and Freeman, William T.},
  booktitle   = {Computer Vision -- ECCV 2024},
  year        = {2024},
  pages       = {388--406},
  doi         = {10.1007/978-3-031-72627-9_22},
}

@inproceedings{cai2024gic,
  title       = {{GIC}: Gaussian-Informed Continuum for Physical Property Identification and Simulation},
  author      = {Cai, Junhao and Yang, Yuji and Yuan, Weihao and He, Yisheng and Dong, Zilong and Bo, Liefeng and Cheng, Hui and Chen, Qifeng},
  booktitle   = {Advances in Neural Information Processing Systems},
  year        = {2024},
  volume      = {37},
  pages       = {75035--75063},
  doi         = {10.52202/079017-2388},
}

@inproceedings{liu2025physflow,
  title       = {Unleashing the Potential of Multi-modal Foundation Models and Video Diffusion for {4D} Dynamic Physical Scene Simulation},
  author      = {Liu, Zhuoman and Ye, Weicai and Luximon, Yan and Wan, Pengfei and Zhang, Di},
  booktitle   = {Proceedings of the IEEE/CVF Conference on Computer Vision and Pattern Recognition (CVPR)},
  year        = {2025},
  pages       = {11016--11025},
  doi         = {10.1109/CVPR52734.2025.01029},
}

@inproceedings{zhao2025physsplat,
  title       = {{PhysSplat}: Efficient Physics Simulation for {3D} Scenes via {MLLM}-Guided Gaussian Splatting},
  author      = {Zhao, Haoyu and Wang, Hao and Zhao, Xingyue and Fei, Hao and Wang, Hongqiu and Long, Chengjiang and Zou, Hua},
  booktitle   = {Proceedings of the IEEE/CVF International Conference on Computer Vision (ICCV)},
  year        = {2025},
  pages       = {5242--5252},
  doi         = {10.1109/ICCV51701.2025.00498},
}

@inproceedings{lin2025omniphysgs,
  title       = {{OmniPhysGS}: {3D} Constitutive Gaussians for General Physics-Based Dynamics Generation},
  author      = {Lin, Yuchen and Lin, Chenguo and Xu, Jianjin and Mu, Yadong},
  booktitle   = {International Conference on Learning Representations},
  year        = {2025},
  pages       = {70160--70187},
  volume      = {2025},
}

@inproceedings{chen2025physgen3d,
  title       = {{PhysGen3D}: Crafting a Miniature Interactive World from a Single Image},
  author      = {Chen, Boyuan and Jiang, Hanxiao and Liu, Shaowei and Gupta, Saurabh and Li, Yunzhu and Zhao, Hao and Wang, Shenlong},
  booktitle   = {Proceedings of the IEEE/CVF Conference on Computer Vision and Pattern Recognition (CVPR)},
  year        = {2025},
  pages       = {6178--6189},
  doi         = {10.1109/CVPR52734.2025.00579},
}

@inproceedings{li2025wonderplay,
  title       = {{WonderPlay}: Dynamic {3D} Scene Generation from a Single Image and Actions},
  author      = {Li, Zizhang and Yu, Hong-Xing and Liu, Wei and Yang, Yin and Herrmann, Charles and Wetzstein, Gordon and Wu, Jiajun},
  booktitle   = {Proceedings of the IEEE/CVF International Conference on Computer Vision (ICCV)},
  year        = {2025},
  pages       = {9080--9090},
  doi         = {10.1109/ICCV51701.2025.00849},
}

@inproceedings{lv2026physgm,
  title       = {{PhysGM}: Large Physical Gaussian Model for Feed-Forward {4D} Synthesis},
  author      = {Lv, Chunji and Chen, Zequn and Di, Donglin and Zhang, Weinan and Li, Hao and Chen, Wei and Lei, Yinjie and Li, Changsheng},
  booktitle   = {Proceedings of the IEEE/CVF Conference on Computer Vision and Pattern Recognition (CVPR)},
  year        = {2026},
  pages       = {29855--29865},
}

@article{chen2026empm,
  title       = {{EMPM}: Embodied {MPM} for Modeling and Simulation of Deformable Objects},
  author      = {Chen, Yunuo and Hu, Yafei and Sun, Lingfeng and Kusnur, Tushar and Herlant, Laura and Jiang, Chenfanfu},
  journal     = {IEEE Robotics and Automation Letters},
  volume      = {11},
  number      = {4},
  pages       = {4179--4186},
  year        = {2026},
  doi         = {10.1109/LRA.2026.3664610},
}

@inproceedings{li2026ngff,
  title       = {Learning Physics-Grounded {4D} Dynamics with Neural Gaussian Force Fields},
  author      = {Li, Shiqian and Shen, Ruihong and Ni, Junfeng and Pan, Chang and Zhang, Chi and Zhu, Yixin},
  booktitle   = {International Conference on Learning Representations},
  year        = {2026},
  pages       = {81165--81207},
  volume      = {2026},
}

@inproceedings{zhan2026perpetualwonder,
  title       = {{PerpetualWonder}: Long-Horizon Action-Conditioned {4D} Scene Generation},
  author      = {Zhan, Jiahao and Li, Zizhang and Yu, Hong-Xing and Wu, Jiajun},
  year        = {2026},
  booktitle   = {Proceedings of the IEEE/CVF Conference on Computer Vision and Pattern Recognition (CVPR)},
  pages       = {25370--25380},
}

@article{lu2026phys4d,
  title       = {{Phys4D}: Fine-Grained Physics-Consistent {4D} Modeling from Video Diffusion},
  author      = {Lu, Haoran and Wu, Shang and Liu, Songling and Zhang, Jianshu and Su, Maojiang and Ye, Guo and Xu, Chenwei and Lu, Lie and Maneriker, Pranav and Du, Fan and Wang, Zhaoran and Liu, Han},
  journal     = {arXiv preprint arXiv:2603.03485},
  year        = {2026},
  doi         = {10.48550/arXiv.2603.03485},
}

@article{liu2026realwonder,
  title       = {{RealWonder}: Real-Time Physical Action-Conditioned Video Generation},
  author      = {Liu, Wei and Chen, Ziyu and Li, Zizhang and Wang, Yue and Yu, Hong-Xing and Wu, Jiajun},
  journal     = {arXiv preprint arXiv:2603.05449},
  year        = {2026},
  doi         = {10.48550/arXiv.2603.05449},
}

@inproceedings{yu2024wonderworld,
  title       = {{WonderWorld}: Interactive {3D} Scene Generation from a Single Image},
  author      = {Yu, Hong-Xing and Duan, Haoyi and Herrmann, Charles and Freeman, William T. and Wu, Jiajun},
  booktitle   = {Proceedings of the IEEE/CVF Conference on Computer Vision and Pattern Recognition (CVPR)},
  year        = {2025},
  pages       = {5916--5926},
  doi         = {10.1109/CVPR52734.2025.00555},
}

@article{ai2025dynamicsreview,
  title       = {A Review of Learning-Based Dynamics Models for Robotic Manipulation},
  author      = {Ai, Bo and Tian, Stephen and Shi, Haochen and Wang, Yixuan and Pfaff, Tobias and Tan, Cheston and Christensen, Henrik I. and Su, Hao and Wu, Jiajun and Li, Yunzhu},
  journal     = {Science Robotics},
  volume      = {10},
  number      = {106},
  pages       = {eadt1497},
  year        = {2025},
  doi         = {10.1126/scirobotics.adt1497},
}

@inproceedings{zhang2024adaptigraph,
  title       = {{AdaptiGraph}: Material-Adaptive Graph-Based Neural Dynamics for Robotic Manipulation},
  author      = {Zhang, Kaifeng and Li, Baoyu and Hauser, Kris and Li, Yunzhu},
  booktitle   = {Proceedings of Robotics: Science and Systems},
  year        = {2024},
  doi         = {10.15607/RSS.2024.XX.010},
}

@article{hu2018mlsmpm,
  title       = {A Moving Least Squares Material Point Method with Displacement Discontinuity and Two-Way Rigid Body Coupling},
  author      = {Hu, Yuanming and Fang, Yu and Ge, Ziheng and Qu, Ziyin and Zhu, Yixin and Pradhana, Andre and Jiang, Chenfanfu},
  journal     = {ACM Transactions on Graphics},
  volume      = {37},
  number      = {4},
  year        = {2018},
  doi         = {10.1145/3197517.3201293},
  pages       = {1--14},
}

@inproceedings{yang2025cogvideox,
  title       = {{CogVideoX}: Text-to-Video Diffusion Models with An Expert Transformer},
  author      = {Yang, Zhuoyi and Teng, Jiayan and Zheng, Wendi and Ding, Ming and Huang, Shiyu and Xu, Jiazheng and Yang, Yuanming and Hong, Wenyi and Zhang, Xiaohan and Feng, Guanyu and Yin, Da and Zhang, Yuxuan and Wang, Weihan and Cheng, Yean and Bin, Xu and Gu, Xiaotao and Dong, Yuxiao and Tang, Jie},
  booktitle   = {International Conference on Learning Representations},
  year        = {2025},
  pages       = {83048--83077},
  volume      = {2025},
}

@inproceedings{zhang2025tora,
  title       = {{Tora}: Trajectory-oriented Diffusion Transformer for Video Generation},
  author      = {Zhang, Zhenghao and Liao, Junchao and Li, Menghao and Dai, Zuozhuo and Qiu, Bingxue and Zhu, Siyu and Qin, Long and Wang, Weizhi},
  booktitle   = {Proceedings of the IEEE/CVF Conference on Computer Vision and Pattern Recognition (CVPR)},
  year        = {2025},
  pages       = {2063--2073},
  doi         = {10.1109/CVPR52734.2025.00198},
}

@inproceedings{gillman2025forceprompt,
  title       = {Force Prompting: Video Generation Models Can Learn and Generalize Physics-based Control Signals},
  author      = {Gillman, Nate and Herrmann, Charles and Freeman, Michael and Aggarwal, Daksh and Luo, Evan and Sun, Deqing and Sun, Chen},
  booktitle   = {Advances in Neural Information Processing Systems},
  year        = {2025},
  volume      = {38, Main Conference},
  pages       = {103174--103201},
  doi         = {10.52202/085713-3448},
}

@inproceedings{gillman2026goalforce,
  title       = {Goal Force: Teaching Video Models To Accomplish Physics-Conditioned Goals},
  author      = {Gillman, Nate and Zhou, Yinghua and Tang, Zitian and Luo, Evan and Chakravarthy, Arjan and Aggarwal, Daksh and Freeman, Michael and Sun, Chen},
  year        = {2026},
  booktitle   = {Proceedings of the IEEE/CVF Conference on Computer Vision and Pattern Recognition (CVPR)},
  pages       = {20077--20087},
}

@inproceedings{karaev2024cotracker3,
  title       = {{CoTracker3}: Simpler and Better Point Tracking by Pseudo-Labelling Real Videos},
  author      = {Karaev, Nikita and Makarov, Yuri and Wang, Jianyuan and Neverova, Natalia and Vedaldi, Andrea and Rupprecht, Christian},
  year        = {2025},
  booktitle   = {Proceedings of the IEEE/CVF International Conference on Computer Vision (ICCV)},
  pages       = {6013--6022},
  doi         = {10.1109/ICCV51701.2025.00568},
}

@inproceedings{lin2026da3,
  title       = {{Depth Anything 3}: Recovering the Visual Space from Any Views},
  author      = {Lin, Haotong and Chen, Sili and Liew, Jun Hao and Chen, Donny Y. and Li, Zhenyu and Zhao, Yang and Peng, Sida and Guo, Hengkai and Zhou, Xiaowei and Shi, Guang and Feng, Jiashi and Kang, Bingyi},
  booktitle   = {International Conference on Learning Representations},
  year        = {2026},
  pages       = {141261--141285},
  volume      = {2026},
}

@inproceedings{wang2025moge2,
  title       = {{MoGe-2}: Accurate Monocular Geometry with Metric Scale and Sharp Details},
  author      = {Wang, Ruicheng and Xu, Sicheng and Dong, Yue and Deng, Yu and Xiang, Jianfeng and Lv, Zelong and Sun, Guangzhong and Tong, Xin and Yang, Jiaolong},
  booktitle   = {Advances in Neural Information Processing Systems},
  year        = {2025},
  volume      = {38, Main Conference},
  pages       = {35928--35959},
  doi         = {10.52202/085713-1207},
}

@inproceedings{xiang2024trellis,
  title       = {Structured {3D} Latents for Scalable and Versatile {3D} Generation},
  author      = {Xiang, Jianfeng and Lv, Zelong and Xu, Sicheng and Deng, Yu and Wang, Ruicheng and Zhang, Bowen and Chen, Dong and Tong, Xin and Yang, Jiaolong},
  booktitle   = {Proceedings of the IEEE/CVF Conference on Computer Vision and Pattern Recognition (CVPR)},
  year        = {2025},
  pages       = {21469--21480},
  doi         = {10.1109/CVPR52734.2025.02000},
}

@inproceedings{sarlin2020superglue,
  title       = {{SuperGlue}: Learning Feature Matching with Graph Neural Networks},
  author      = {Sarlin, Paul-Edouard and DeTone, Daniel and Malisiewicz, Tomasz and Rabinovich, Andrew},
  booktitle   = {Proceedings of the IEEE/CVF Conference on Computer Vision and Pattern Recognition (CVPR)},
  year        = {2020},
  pages       = {4937--4946},
  doi         = {10.1109/CVPR42600.2020.00499},
}

@inproceedings{huang2024scgs,
  title       = {{SC-GS}: Sparse-Controlled Gaussian Splatting for Editable Dynamic Scenes},
  author      = {Huang, Yi-Hua and Sun, Yang-Tian and Yang, Ziyi and Lyu, Xiaoyang and Cao, Yan-Pei and Qi, Xiaojuan},
  booktitle   = {Proceedings of the IEEE/CVF Conference on Computer Vision and Pattern Recognition (CVPR)},
  year        = {2024},
  pages       = {4220--4230},
  doi         = {10.1109/CVPR52733.2024.00404},
}

@article{sumner2007embedded,
  title       = {Embedded Deformation for Shape Manipulation},
  author      = {Sumner, Robert W. and Schmid, Johannes and Pauly, Mark},
  journal     = {ACM Transactions on Graphics},
  volume      = {26},
  number      = {3},
  year        = {2007},
  doi         = {10.1145/1276377.1276478},
  pages       = {80},
}

@misc{ren2024grounded,
      title={Grounded SAM: Assembling Open-World Models for Diverse Visual Tasks}, 
      author={Tianhe Ren and Shilong Liu and Ailing Zeng and Jing Lin and Kunchang Li and He Cao and Jiayu Chen and Xinyu Huang and Yukang Chen and Feng Yan and Zhaoyang Zeng and Hao Zhang and Feng Li and Jie Yang and Hongyang Li and Qing Jiang and Lei Zhang},
      year={2024},
      eprint={2401.14159},
      archivePrefix={arXiv},
      primaryClass={cs.CV}
}

@inproceedings{zhang2025real2simeval,
  title       = {Real-to-Sim Robot Policy Evaluation with {Gaussian} Splatting Simulation of Soft-Body Interactions},
  author      = {Zhang, Kaifeng and Sha, Shuo and Jiang, Hanxiao and Loper, Matthew and Song, Hyunjong and Cai, Guangyan and Xu, Zhuo and Hu, Xiaochen and Zheng, Changxi and Li, Yunzhu},
  year        = {2026},
  booktitle   = {2026 IEEE International Conference on Robotics and Automation (ICRA)},
  pages       = {22356--22363},
  doi         = {10.1109/icra57385.2026.11696757},
}

@incollection{hansen2006cma,
  title       = {The {CMA} Evolution Strategy: A Comparing Review},
  author      = {Hansen, Nikolaus},
  booktitle   = {Towards a New Evolutionary Computation: Advances in the Estimation of Distribution Algorithms},
  editor      = {Lozano, Jose A. and Larra\~naga, Pedro and Inza, I\~naki and Bengoetxea, Endika},
  series      = {Studies in Fuzziness and Soft Computing},
  volume      = {192},
  publisher   = {Springer},
  address     = {Berlin, Heidelberg},
  pages       = {75--102},
  year        = {2006},
  doi         = {10.1007/3-540-32494-1_4},
}

@article{kerbl2023gaussian,
  title       = {{3D Gaussian Splatting} for Real-Time Radiance Field Rendering},
  author      = {Kerbl, Bernhard and Kopanas, Georgios and Leimk{\"u}hler, Thomas and Drettakis, George},
  journal     = {ACM Transactions on Graphics},
  volume      = {42},
  number      = {4},
  year        = {2023},
  doi         = {10.1145/3592433},
  pages       = {1--14},
}

@inproceedings{wu2025sport,
  title       = {Stimulating Imagination: Towards General-purpose ``Something Something Placement''},
  author      = {Wu, Jianyang and Gu, Jie and Ma, Xiaokang and Qiu, Fangzhou and Tang, Chu and Chen, Jingmin},
  booktitle   = {2025 IEEE/RSJ International Conference on Intelligent Robots and Systems (IROS)},
  year        = {2025},
  doi         = {10.1109/IROS60139.2025.11245844},
  pages       = {701--707},
}

@inproceedings{wang2026vggtomega,
  author      = {Wang, Jianyuan and Chen, Minghao and Zhang, Shangzhan and Karaev, Nikita and Sch\"onberger, Johannes and Labatut, Patrick and Bojanowski, Piotr and Novotny, David and Vedaldi, Andrea and Rupprecht, Christian},
  title       = {{VGGT-$\Omega$}},
  booktitle   = {Proceedings of the IEEE/CVF Conference on Computer Vision and Pattern Recognition (CVPR)},
  month       = {June},
  year        = {2026},
  pages       = {21486--21499},
}

@inproceedings{li2025orderedkeypoints,
  title       = {Learning-based Keypoints Detection with Topological Order on Deformable Linear Objects from Incomplete Point Clouds},
  author      = {Li, Can and Liu, Jingyang and Sun, Lei},
  booktitle   = {2025 IEEE/RSJ International Conference on Intelligent Robots and Systems (IROS)},
  year        = {2025},
  pages       = {650--656},
  doi         = {10.1109/iros60139.2025.11246954},
}

@InProceedings{zhao2025masiv,
    author    = {Zhao, Yizhou and Chen, Haoyu and Liu, Chunjiang and Li, Zhenyang and Herrmann, Charles and Hur, Junhwa and Li, Yinxiao and Yang, Ming-Hsuan and Raj, Bhiksha and Xu, Min},
    title     = {Toward Material-Agnostic System Identification from Videos},
    booktitle = {Proceedings of the IEEE/CVF International Conference on Computer Vision (ICCV)},
    month     = {October},
    year      = {2025},
    pages     = {5944-5956}
}

@inproceedings{cao2024neuma,
 author = {Cao, Junyi and Guan, Shanyan and Ge, Yanhao and Li, Wei and Yang, Xiaokang and Ma, Chao},
 booktitle = {Advances in Neural Information Processing Systems},
 doi = {10.52202/079017-2097},
 pages = {65643--65669},
 title = {NeuMA: Neural Material Adaptor for Visual Grounding of Intrinsic Dynamics},
 volume = {37},
 year = {2024}
}

@inproceedings{lin2026visionlaw,
 author = {Lin, Jiajing and Jiang, Shu and Zeng, Qingyuan and Wang, Zhenzhong and Jiang, Min},
 booktitle = {International Conference on Learning Representations},
 pages = {96128--96165},
 title = {VisionLaw: Inferring Interpretable Intrinsic Dynamics from Visual Observations via Bilevel Optimization},
 volume = {2026},
 year = {2026}
}

@InProceedings{ma2023nclaw,
  title = 	 {Learning Neural Constitutive Laws from Motion Observations for Generalizable {PDE} Dynamics},
  author =       {Ma, Pingchuan and Chen, Peter Yichen and Deng, Bolei and Tenenbaum, Joshua B. and Du, Tao and Gan, Chuang and Matusik, Wojciech},
  booktitle = 	 {Proceedings of the 40th International Conference on Machine Learning},
  pages = 	 {23279--23300},
  year = 	 {2023},
  volume = 	 {202},
  series = 	 {Proceedings of Machine Learning Research},
}
\bibliographystyle{plainnat}

\newpage
\appendix
\section{Appendix}
\label{sec:appendix}

\subsection{Online Interactive Playground}
\label{app:online_playground}

\begin{figure}[h]
    \centering
    \includegraphics[width=\linewidth]{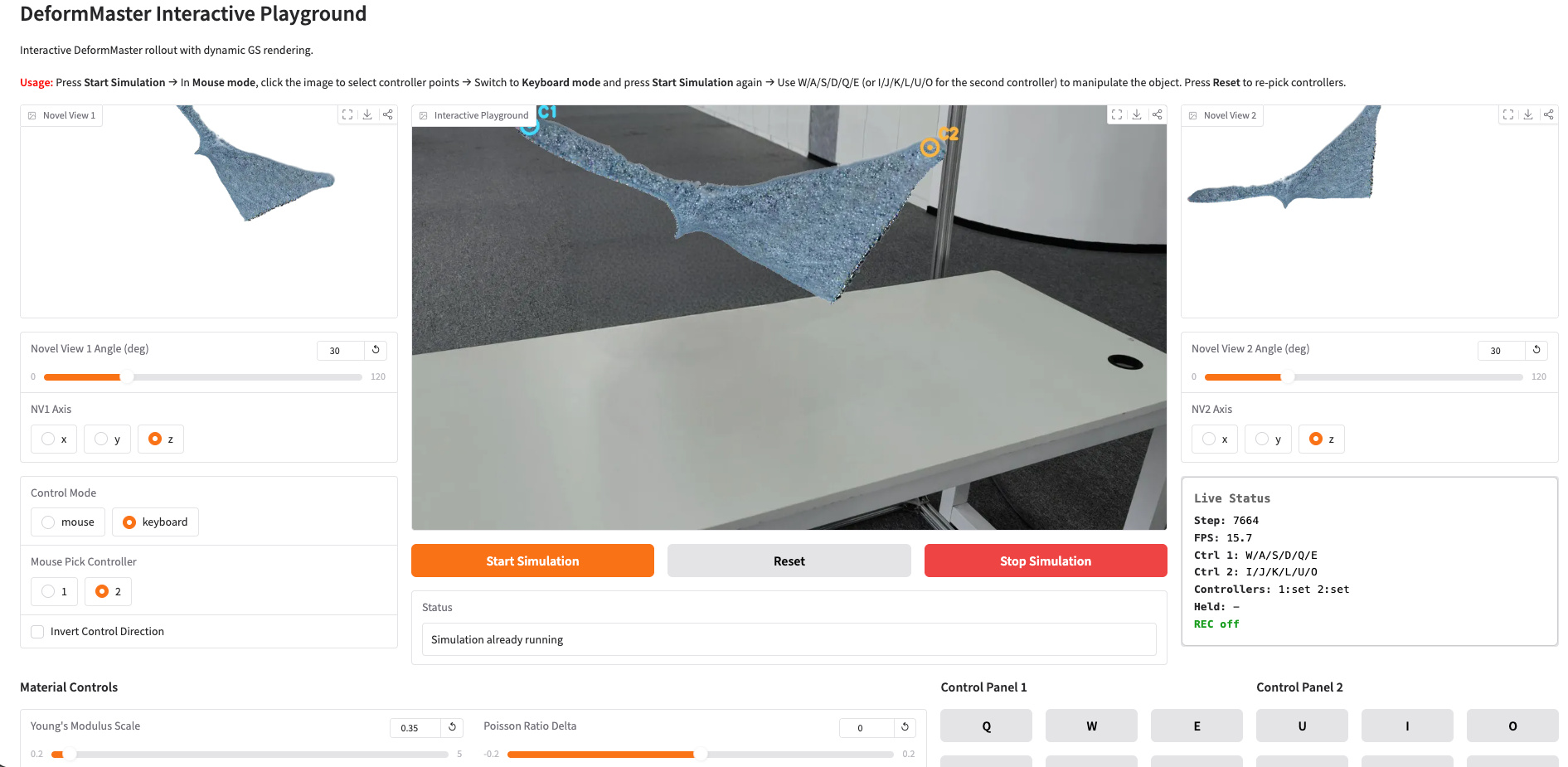}
    \caption{Online interactive playground for DeformMaster. The interface supports interaction-point selection, keyboard-based object manipulation, material-parameter adjustment, and synchronized novel-view rendering during online rollout. The live status panel reports the online rollout speed, showing real-time interaction at over $15$\,fps in this example.}
    \label{fig:online_playground}
\end{figure}

We provide an online interactive playground for DeformMaster, where users can select interactive points, adjust material parameters, manipulate the deformable object through keyboard inputs, and inspect synchronized novel-view renderings during rollout.
\Cref{fig:online_playground} shows the playground interface, including the central interactive view, two novel-view renderers, controller status, material parameter adjustment, and keyboard control panels.

\subsection{Downstream Applications}
\label{app:downstream_applications}

\begin{figure}[h]
    \centering
    \includegraphics[width=\linewidth]{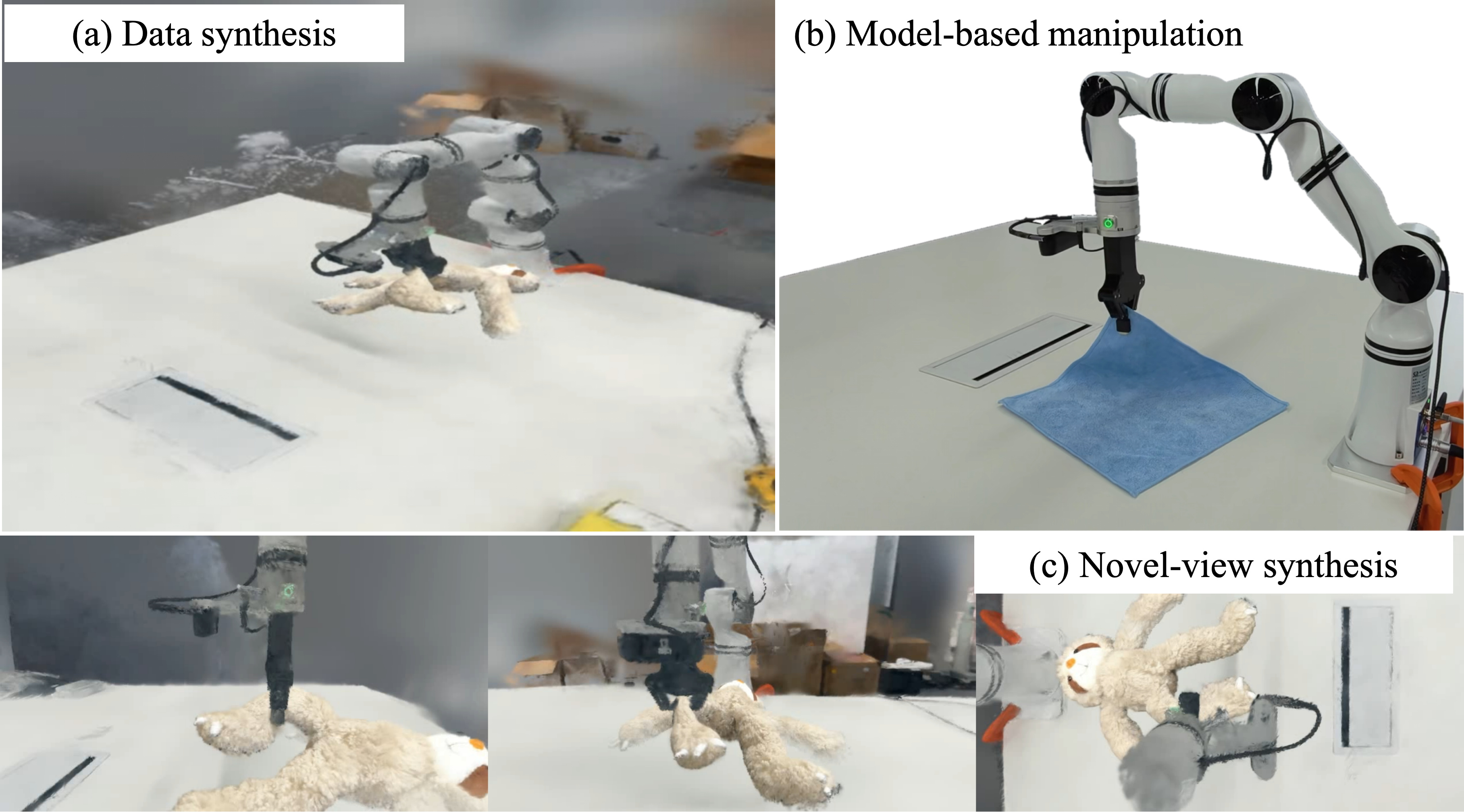}
    \caption{Applications of DeformMaster. (a) DeformMaster leverages interactive manipulation to synthesize additional deformable-object data beyond the observed video. (b) DeformMaster enables model predictive control for deformable-object manipulation. (c) It supports convenient visualization and dynamic novel-view image synthesis.}
    \label{fig:applications}
\end{figure}

The interactive world model DeformMaster enables a range of embodied downstream tasks. As shown in \Cref{fig:applications}, its interaction capability can be used to synthesize additional deformable-object data, while the model supports model-predictive-control-based robotic manipulation. The resulting 4D representation also provides convenient visualization and novel-view image synthesis.

\subsection{Physics Preliminaries}
\label{app:preliminaries}
\label{app:mpm_config}

MPM \citep{hu2018mlsmpm} represents a deformable object with particles carrying position, velocity, and deformation information. Each substep transfers particle quantities to a background grid, updates grid velocities under internal and external forces, and transfers the updated motion back to particles.

A constitutive law maps deformation to stress, determining how the material resists deformation. In DeformMaster, MoCE combines constitutive experts with spatially varying weights to supply stresses for the MPM update (\cref{eq:moe}).

Tracked hand or actuator positions and velocities define moving anchors. DCA couples these anchors to nearby object particles, converting relative position and velocity errors into distributed accelerations that drive the simulated motion (\cref{eq:dca}).

\Cref{tab:mpm_config} lists the MPM substrate parameters used throughout training and rollout. Volumetric object interior particle filling follows PhysGaussian \citep{xie2024physgaussian}. One video-frame transition applies $N_\text{sub} = \Delta t / \delta t$ MPM substeps. Per-category numeric overrides are noted; values without an override are shared.

\begin{table}[h]
    \centering
    \caption{MPM solver configuration. \emph{linear / planar / volumetric} columns indicate per-category overrides where they differ from the default.}
    \label{tab:mpm_config}
    \small
    \begin{tabular}{l l l}
        \toprule
        Group & Symbol / name & Value \\
        \midrule
        \multicolumn{3}{l}{\emph{Discretisation}} \\
        Maximum particles per scene & $N_{\max}$                    & $10^{5}$ \\
        Background grid         & $G$                               & $32^3$ \\
        Kernel                  & cubic B-spline                    & 27-node support \\
        Substep                 & $\delta t$                        & $8\!\times\!10^{-4}$\,s \\
        Frame interval          & $\Delta t$                        & $1/30$\,s \\
        Substeps per frame      & $N_\text{sub} = \Delta t / \delta t$  & $42$ \\
        \midrule
        \multicolumn{3}{l}{\emph{Forces \& boundary}} \\
        Gravity                 & $\mathbf{g}$                      & $(0,\,0,\,-9.8)$\,m/s$^2$ \\
        Particle damping        & -                               & $20.0$ (planar/linear), $5.0$ (volumetric) \\
        Grid velocity damping   & multiplicative per substep        & $0.999$ \\
        Position clip           & $[\mathbf{x}_\text{min}, \mathbf{x}_\text{max}]$ & $[0.05,\,0.95]^3$ (in shifted unit cube) \\
        Floor margin            & -                               & $0.0$ (planar/linear), $0.05$ (volumetric) \\
        \midrule
        \multicolumn{3}{l}{\emph{Constitutive (per-particle, log-sigmoid bounded)}} \\
        Young's modulus bounds  & $E$                               & \shortstack[l]{$[10^4, 10^6]$\,Pa (planar) \\ $[10^5, 10^7]$\,Pa (volumetric)} \\
        Poisson ratio bounds    & $\nu$                             & \shortstack[l]{$[0,\,0.35]$ (planar) \\ $[0,\,0.45]$ (volumetric)} \\
        SVD clamp               & on $\mathbf{F}$ singular values   & $[1/2.0,\,2.0]$ \\
        Active experts          & MoCE (\cref{app:moe_patch_param})    & \{NH, Cor., StVK\} \\
        \midrule
        \multicolumn{3}{l}{\emph{Differentiation}} \\
        Backend                 & NVIDIA Warp         & - \\
        Truncated-BPTT window   & last-$W$ substeps recorded        & $W = 20$ \\
        Per-parameter clip      & on physics gradients              & $1.0$ \\
        Global gradient clip    & on physics parameters             & $10.0$ \\
        \bottomrule
    \end{tabular}
\end{table}

\subsection{Neural Residual Architecture}
\label{app:residual_pgnd_arch}

The main text describes neural particle-grid residual and gives the encoder equation in \cref{eq:residual_pgnd_encoder}. Here we provide the decoder and particle-grid interpolation equations that complete the neural residual $\mathcal{R}_\phi$.

The encoder $E_\phi$ is a PointNet \citep{qi2017pointnet} whose inputs are the post-MPM position $\tilde{\mathbf{x}}_p^{t+1}$ and velocity $\tilde{\mathbf{v}}_p^{t+1}$ of particle $p$ (components of $\tilde s_{t+1}$), the displacement $\tilde{\mathbf{x}}_p^{t+1} - \mathbf{x}_p^{t}$ accumulated by MPM during the current frame, and the kinematic history $h_t$ of \cref{eq:hybrid_residual}. The decoder maps each grid node coordinate $\mathbf{x}_g$ to a node-level correction $\mathbf{u}_g$ using a coordinate-conditioned MLP,
\begin{equation}
    \mathbf{u}_g \;=\; \alpha \cdot \tanh\!\big(D_\phi(\mathbf{f}_p,\, \gamma(\mathbf{x}_g))\big),
    \label{eq:residual_pgnd_decoder}
\end{equation}
where $D_\phi$ is the MLP, $\gamma(\cdot)$ is Fourier positional encoding of the node coordinate, and $\alpha$ is a scalar that bounds the magnitude of the per-particle correction. Finally, the per-particle correction is the B-spline-weighted sum over the surrounding grid nodes,
\begin{equation}
    \Delta\mathbf{v}_p \;=\; \sum_{g \,\in\, \mathcal{N}_p} w_{p,g}\,\mathbf{u}_g,
    \label{eq:residual_pgnd_interp}
\end{equation}
where $\mathcal{N}_p$ is the set of grid nodes supporting particle $p$ under MPM's B-spline interpolation, and $w_{p,g}$ are the same interpolation weights used by MPM for P2G/G2P transfers. This mirrors MPM's particle-grid hybrid at the level of grid discretization and particle-grid transfer: the encoder extracts particle features from the point set, while the decoder predicts local node-wise residuals on each particle's supporting grid neighborhood before interpolating them back to particles. Concrete network widths and training hyperparameters are listed in \cref{tab:hparams_pgnd}.

\begin{table}[h]
    \centering
    \caption{Neural Residual configuration. Values are shared across material categories unless otherwise noted; MPM solver settings (substep, grid, etc.) are listed separately in \cref{tab:mpm_config}.}
    \label{tab:hparams_pgnd}
    \small
    \begin{tabular}{l l l}
        \toprule
        Group & Symbol / name & Value \\
        \midrule
        \multicolumn{3}{l}{\emph{Inputs}} \\
        Per-particle channels   & $9 + 6H$                          & $27$ (with $H{=}3$) \\
        \quad ground fields     & pos, vel, frame displacement      & $9$ \\
        \quad history           & $H$ frames of (pos, vel)          & $6H$ \\
        Centring                & subtract particle-cloud centroid  & yes \\
        \midrule
        \multicolumn{3}{l}{\emph{PointNet encoder (Lagrangian)}} \\
        Conv1D widths           & $(9{+}6H) \to 64 \to 128 \to 64$  & - \\
        Normalisation           & GroupNorm                         & - \\
        Output                  & per-particle feature              & $64$-dim \\
        \midrule
        \multicolumn{3}{l}{\emph{Neural-field decoder (Eulerian)}} \\
        Query nodes / particle  & MPM B-spline support              & $27$ (cubic kernel) \\
        Decoder grid            & co-located with MPM grid          & see \cref{tab:mpm_config} \\
        Positional encoding     & Fourier bands $L$                 & $6$ ($\dim 15$) \\
        MLP                     & layers $\times$ width             & $4 \times 128$ \\
        Skip connection         & every $4$ layers                  & yes \\
        Output activation       & $\tanh \cdot \alpha$              & $\alpha = 1.0$\,m/s \\
        Particle interpolation  & B-spline weights (P2G/G2P)        & 27-node \\
        Residual cadence        & once per frame (post all substeps)& - \\
        \midrule
        \multicolumn{3}{l}{\emph{Optimisation}} \\
        Optimiser               & Adam                              & $\beta_1{=}0.9,\, \beta_2{=}0.999$ \\
        Learning rate           & $\eta_\phi$                       & $5\!\times\!10^{-3}$ \\
        Gradient norm clip      & -                               & $5.0$ \\
        Residual regulariser    & $\lambda_\text{reg}\,\|\Delta\mathbf{v}\|^2$ & $\lambda_\text{reg} = 10^{-3}$ \\
        Differentiability       & MPM via Warp autodiff             & end-to-end \\
        Truncated-BPTT window   & MPM tape length (in substeps)     & see \cref{tab:mpm_config} \\
        \bottomrule
    \end{tabular}
\end{table}

\subsection{MoCE Parameterization}
\label{app:moe_patch_param}

The main text defines the constitutive mixture at the particle level in \cref{eq:moe}. The mixture is taken over three canonical hyperelastic experts: \textbf{Neo-Hookean (NH)}, \textbf{fixed Corotated (Cor)}, and \textbf{St.~Venant--Kirchhoff (StVK)}, each producing a first Piola--Kirchhoff stress $\mathbf{P}_k(\mathbf{F}_p; E_p, \nu_p)$ from the deformation gradient under shared Young's modulus and Poisson's ratio. In the implementation, the learnable physical parameters are stored on a persistent set of material patches and interpolated to particles before each MPM rollout.

Patch centers are sampled once from the first-frame geometry using farthest-point sampling. Each simulation particle $p$ is assigned its three nearest patch anchors $\mathcal{A}(p)$, and the normalized inverse-distance weights
\begin{equation}
    \beta_{p,c}
    =
    \frac{1 / d(\mathbf{x}_p^0,\mathbf{a}_c)}
         {\sum_{c' \in \mathcal{A}(p)} 1 / d(\mathbf{x}_p^0,\mathbf{a}_{c'})},
    \qquad c \in \mathcal{A}(p),
    \label{eq:moe_patch_interp_weights}
\end{equation}
are kept fixed so that patch parameters remain tied to the same material regions during optimization. The expert weights are represented by patch logits $\ell_{k,c}$ and converted to patch probabilities by
\begin{equation}
    q_{k,c} = \mathrm{softmax}_k(\ell_{k,c}).
    \label{eq:moe_patch_softmax}
\end{equation}
The particle-level mixture weights used in \cref{eq:moe} are then
\begin{equation}
    w_{k,p} = \sum_{c \in \mathcal{A}(p)} \beta_{p,c}\,q_{k,c}.
    \label{eq:moe_weights}
\end{equation}
The same patch-to-particle interpolation is applied to the bounded Young's modulus and Poisson's ratio parameters before converting them to Lam\'e parameters for the MPM stress computation.

\subsection{Video Preprocessing Pipeline}
\label{app:video_preprocess}

Each input clip is processed by a fixed cascade of foundation models that converts raw RGB(\nobreakdash-D) frames into the data structures consumed by the trainer (object masks, dense 3D point tracks for the object and the actuator, a first-frame canonical mesh, and calibrated camera intrinsics/extrinsics). The cascade is run once offline per clip and shared across all training stages.

\paragraph{Object segmentation and tracking.}
Object masks are produced by Grounded\nobreakdash-SAM\nobreakdash-2 \citep{ren2024grounded}: a text prompt naming the target object (e.g.\ ``cloth'', ``rope'', ``stuffed toy'') is fed to Grounding~DINO for open-vocabulary detection on the first frame, the resulting bounding box is passed to SAM~2 to obtain a high-fidelity initial mask, and SAM~2's video-mode tracker propagates the mask through all $T$ frames. The same pipeline is run with prompt ``hand'' to obtain a per-frame controller mask, used both to exclude hand pixels from the photometric loss and to anchor actuator positions.

\paragraph{Dense 2D--3D tracking.}
Within the object mask we sample a dense set of query points on the first frame (farthest-point sampling on the masked region) and propagate them across time with CoTracker3 \citep{karaev2024cotracker3}, whose pseudo-label\nobreakdash-trained design is robust to occlusion and partial out-of-view. The resulting 2D tracks are unprojected to 3D using the per-frame depth from the next step, producing the dense $(T, N, 3)$ point trajectory that drives the Chamfer/Track losses in $\mathcal{L}_\text{dynamics}$.

\paragraph{Monocular geometry (uncalibrated single-camera setting).}
For monocular sequences we estimate per-frame depth and camera pose with VGGT-Omega \citep{wang2026vggtomega} or Depth~Anything~3 \citep{lin2026da3}. We resolve global scale by fitting a per-clip scale factor that aligns its predictions with the metric point map produced by MoGe\nobreakdash-2 \citep{wang2025moge2} on the masked object. We apply the estimated scale consistently to the predicted depth maps and camera translations, aligning the reconstruction with MoGe\nobreakdash-2's metric-scale geometric prior.

\paragraph{Shape generative prior.}
To initialize particles in regions occluded at $t{=}0$, we use TRELLIS \citep{xiang2024trellis} to generate a canonical textured mesh from the segmented first frame. The mesh is aligned to the observed first-frame point cloud by rendering candidate mesh views, selecting the best match to the real RGB crop with SuperGlue \citep{sarlin2020superglue} keypoint matching, solving the 6-DoF pose with PnP, and then refining it with an ARAP regularizer. The aligned mesh is volumetrically sampled together to populate the MPM particles.

\subsection{Training Details}
\label{app:training}

\paragraph{Stage 1--2 alternating optimization.}

The interactive physics-neural dynamics is implemented with a mixed NVIDIA Warp/PyTorch backend: the MPM simulator runs in Warp, while the neural residual is implemented in PyTorch. The MPM step is differentiable end-to-end via NVIDIA Warp's autodiff backend, so the residual is optimised jointly with the physics parameters rather than in two stages.

Stages 1 and 2 are coupled: DCA gains $(k_p,k_d)$ and material fields are mutually dependent through the simulator rollout. We therefore alternate between gradient-based dynamics training and gradient-free gain selection. We first optimize the material fields and neural residual with default gains, then run CMA-ES gain selection using the warm-started dynamics, and finally continue optimizing the material fields and neural residual with the selected gains. 

We add a small $\ell_2$ penalty $\lambda_\text{reg}\|\Delta\mathbf{v}\|^2$ to the training loss to discourage unnecessary correction; the value of $\lambda_\text{reg}$ is given in \cref{tab:hparams_pgnd}, and the truncated-BPTT horizon over which gradients are propagated is set by the MPM tape window in \cref{tab:mpm_config}.

\paragraph{RGB-guided refinement in Training Stage~3.}
Stage~3 has two steps. We first optimize a first-frame Gaussian Splatting representation $\mathcal{G}_0$ with appearance supervision, then freeze $\mathcal{G}_0$ and resume from the Stage~2 dynamics checkpoint for $30$ additional iterations. During this refinement step, $\mathcal{G}_0$ is deformed by linear blend skinning along the predicted particle trajectory $\mathbf{x}_{1:t}$, rasterised through each training-view camera, and compared to the recorded frame using a masked $\ell_1$ RGB loss. Object pixels retain their captured RGB, background pixels are set to the renderer background to penalize Gaussian bleed, and hand pixels are excluded because the renderer contains no hand model. The refinement loss keeps the dynamics terms and adds RGB supervision, $\mathcal{L}_\text{stage3}=\mathcal{L}_\text{dynamics}+\lambda_\text{rgb}\mathcal{L}_\text{rgb}$ with $\lambda_\text{rgb}=10^{-2}$. Because $\mathcal{G}_0$ is deformed from the predicted particle trajectory, the photometric gradient flows through LBS back to particle motion and the neural residual, acting as an RGB-guided constraint on the dynamics rather than an update to the frozen Gaussians.



\subsection{Novel-Action and Dynamic Novel-View Evaluation}
\label{app:novel_action_sequences}

We conduct a controlled same-object cross-action evaluation, using object dynamics learned from one interaction sequence to predict responses to unseen action trajectories from another sequence of the same object, with all learned parameters fixed. The unseen actions involve different contacts, directions, magnitudes, and trajectories. Five same-object groups contain 15 interaction sequences. For example, the zebra group contains ``single\_\allowbreak{}lift\_\allowbreak{}zebra'', ``double\_\allowbreak{}lift\_\allowbreak{}zebra'', and ``double\_\allowbreak{}stretch\_\allowbreak{}zebra''; transferring from ``single\_\allowbreak{}lift\_\allowbreak{}zebra'' to ``double\_\allowbreak{}lift\_\allowbreak{}zebra'' and transferring in the reverse direction count as two separate evaluations. Evaluating all ordered pairs of distinct sequences within the five object groups yields 32 directed source-to-target evaluations per method. The source checkpoint and learned parameters, including the material parameters, DCA gains, and physics-neural dynamics parameters for DeformMaster, remain fixed, and the solver settings are unchanged. The target sequence supplies only the initial geometry and controller trajectory. No target-action training, adaptation, or parameter selection is used; target observations serve only as ground truth on the future-test split.

We also evaluate dynamic novel-view synthesis by rendering predicted future dynamics from unseen viewpoints. \Cref{tab:novel_view} reports appearance metrics on the future-test frames.

\begin{table}[H]
    \centering
    \caption{Quantitative dynamic novel-view synthesis on future-test frames from unseen views.}
    \label{tab:novel_view}
    \small
    \setlength{\tabcolsep}{4pt}
    \begin{tabular}{l ccc}
        \toprule
        Method & PSNR $\uparrow$ & SSIM $\uparrow$ & LPIPS $\downarrow$ \\
        \midrule
        PhysTwin \citep{jiang2025phystwin} & 28.6606 & 0.9693 & 0.0341 \\
        \textbf{DeformMaster (ours)} & \textbf{28.8056} & \textbf{0.9703} & \textbf{0.0319} \\
        \bottomrule
    \end{tabular}
\end{table}

\subsection{Ablation Details}
\label{app:ablation_tables}

\Cref{tab:ablation_residual_overall,tab:ablation_residual,tab:ablation_dca_overall,tab:ablation_dca,tab:ablation_moe_overall,tab:ablation_moe,tab:stage3_ablation_overall,tab:stage3_ablation} provide the full numerical ablations for the residual branch, DCA, MoCE, and Stage~3 RGB-guided dynamics refinement. Each ablation is reported in two views: an \emph{overall} table with all six paper metrics (\cref{tab:ablation_residual_overall,tab:ablation_dca_overall,tab:ablation_moe_overall,tab:stage3_ablation_overall}) and a \emph{per-category} dynamics table grouped by Linear / Planar / Volumetric (\cref{tab:ablation_residual,tab:ablation_dca,tab:ablation_moe,tab:stage3_ablation}).

\subsection{Runtime, Memory, and Scalability}
\label{app:runtime}

\Cref{tab:runtime_breakdown,tab:runtime_scaling,tab:offline_costs} provide the detailed measurements supporting \cref{sec:runtime}. We profile ``double\_\allowbreak{}lift\_\allowbreak{}sloth'' on one NVIDIA A100-SXM4-80GB, using 22,132 physical particles, 102,613 Gaussians, a $32^3$ grid, 42 MPM substeps per frame, and $848\times480$ rendering as the default configuration.

For the component breakdown in \cref{tab:runtime_breakdown}, we use five warmup frames, 50 measured frames, three repetitions, and synchronized timing. Dynamics and appearance are timed independently; the end-to-end latency is their derived sum, and peak GPU memory is measured separately for each. The dynamics-only sweeps in \cref{tab:runtime_scaling} report medians over nine measured repeats, varying one factor at a time. Latency is reported per frame except for the horizon sweep, which reports total rollout time. \Cref{tab:offline_costs} lists preprocessing and one-time training costs separately from online inference.

\subsection{Limitations}
\label{sec:experiments:limitations}

DeformMaster still has several limitations. First, particle-grid simulation is less specialized for very thin planar objects, such as single-layer cloth, than spring-mass systems with fixed topological links. This is reflected by the remaining gap to PhysTwin on planar object Track (\cref{sec:experiments:main}), where explicit spring-mass connectivity provides a strong inductive bias for cloth-like bending and stretching. Second, interaction modeling relies on observed hand motion. When the hand or actuator is severely occluded in the input video, the recovered interaction signal can be incomplete, which in turn degrades the action-conditioned rollout. Third, runtime depends on the number of physics substeps used during simulation. Reducing the substep count of the MPM solver improves real-time performance, whereas high-accuracy deformation simulation often requires more substeps; this creates a trade-off between fidelity and runtime. Finally, the learned material fields and MoCE weights are effective parameters, not uniquely identified physical constants. The model remains instance-specific and does not support zero-shot generalization to arbitrary new objects.

\begin{table}[H]
    \centering
    \caption{Residual ablation: overall test metrics on the 20-case PhysTwin benchmark. Chamfer and Track in units of $10^{-2}$.}
    \label{tab:ablation_residual_overall}
    \small
    \setlength{\tabcolsep}{4pt}
    \begin{tabular}{l ccc ccc}
        \toprule
        \multirow{2}{*}{Variant} & \multicolumn{3}{c}{Future dynamics} & \multicolumn{3}{c}{Future appearance} \\
        \cmidrule(lr){2-4}\cmidrule(lr){5-7}
        & IoU $\uparrow$ & Chamfer $\downarrow$ & Track $\downarrow$ & PSNR $\uparrow$ & SSIM $\uparrow$ & LPIPS $\downarrow$ \\
        \midrule
        \textbf{Ours} (PNPGD) & \textbf{0.748} & \textbf{1.14} & \textbf{2.40} & \textbf{25.41} & \textbf{0.936} & \textbf{0.061} \\
        MLP                   & 0.735 & 1.27 & 2.59 & 25.16 & 0.935 & 0.063 \\
        GNN                   & 0.715 & 1.44 & 2.78 & 24.68 & 0.932 & 0.068 \\
        No residual           & 0.656 & 2.12 & 3.94 & 23.90 & 0.929 & 0.076 \\
        \bottomrule
    \end{tabular}
\end{table}

\begin{table}[H]
    \centering
    \caption{Ablation on the residual branch $\mathcal{R}_\phi$: per-category dynamics on the PhysTwin benchmark. Chamfer and Track in units of $10^{-2}$. \textbf{Ours}: neural particle-grid. \textbf{MLP}: per-particle correction with no spatial structure. \textbf{GNN}: 3-round KNN graph network. \textbf{No residual}: pure MPM rollout.}
    \label{tab:ablation_residual}
    \small
    \setlength{\tabcolsep}{3.5pt}
    \begin{tabular}{l ccc ccc ccc}
        \toprule
        \multirow{2}{*}{Variant} & \multicolumn{3}{c}{Linear ($n{=}3$)} & \multicolumn{3}{c}{Planar ($n{=}9$)} & \multicolumn{3}{c}{Volumetric ($n{=}8$)} \\
        \cmidrule(lr){2-4}\cmidrule(lr){5-7}\cmidrule(lr){8-10}
        & IoU $\uparrow$ & Chamfer $\downarrow$ & Track $\downarrow$ & IoU $\uparrow$ & Chamfer $\downarrow$ & Track $\downarrow$ & IoU $\uparrow$ & Chamfer $\downarrow$ & Track $\downarrow$ \\
        \midrule
        \textbf{Ours} & 0.721 & \textbf{0.50} & \textbf{1.01} & \textbf{0.748} & \textbf{1.30} & \textbf{3.19} & \textbf{0.756} & \textbf{1.19} & \textbf{2.02} \\
        MLP                   & \textbf{0.727} & 0.52 & 1.04 & 0.733 & 1.45 & 3.43 & 0.739 & 1.35 & 2.21 \\
        GNN                   & 0.700 & 0.57 & 1.19 & 0.719 & 1.66 & 3.63 & 0.714 & 1.52 & 2.42 \\
        No residual           & 0.628 & 0.78 & 1.48 & 0.695 & 2.08 & 4.52 & 0.614 & 2.67 & 4.20 \\
        \bottomrule
    \end{tabular}
\end{table}

\begin{table}[H]
    \centering
    \caption{DCA ablation: overall test metrics. Chamfer and Track in units of $10^{-2}$. $^\ddagger$ rigid is averaged over the $11/20$ sequences that survived training (the $8$ Volumetric and $1$ Linear that diverged are excluded).}
    \label{tab:ablation_dca_overall}
    \small
    \setlength{\tabcolsep}{4pt}
    \begin{tabular}{l ccc ccc}
        \toprule
        \multirow{2}{*}{Variant} & \multicolumn{3}{c}{Future dynamics} & \multicolumn{3}{c}{Future appearance} \\
        \cmidrule(lr){2-4}\cmidrule(lr){5-7}
        & IoU $\uparrow$ & Chamfer $\downarrow$ & Track $\downarrow$ & PSNR $\uparrow$ & SSIM $\uparrow$ & LPIPS $\downarrow$ \\
        \midrule
        \textbf{DCA (ours)}  & \textbf{0.748} & \textbf{1.14} & \textbf{2.40} & \textbf{25.41} & \textbf{0.936} & \textbf{0.061} \\
        rigid$^{\ddagger}$   & 0.713 & 1.44 & 3.25 & 24.51 & 0.927 & 0.078 \\
        single actuator      & 0.561 & 3.08 & 6.19 & 22.32 & 0.914 & 0.102 \\
        \bottomrule
    \end{tabular}
\end{table}

\begin{table}[H]
    \centering
    \caption{DCA ablation: per-category dynamics. Chamfer and Track in units of $10^{-2}$. \textbf{rigid} (no compliance): replace compliant coupling with hard Dirichlet boundary. \textbf{single actuator} (no distribution): keep compliance but actuate each particle through a single actuator instead of a distributed contact patch. ``NaN'' marks the $8$ Volumetric sequences that diverged near iteration~$0$ under hard pointwise constraints; $^\dagger$ rigid completes only $2/3$ Linear sequences.}
    \label{tab:ablation_dca}
    \small
    \setlength{\tabcolsep}{3.5pt}
    \begin{tabular}{l ccc ccc ccc}
        \toprule
        \multirow{2}{*}{Variant} & \multicolumn{3}{c}{Linear ($n{=}3$)} & \multicolumn{3}{c}{Planar ($n{=}9$)} & \multicolumn{3}{c}{Volumetric ($n{=}8$)} \\
        \cmidrule(lr){2-4}\cmidrule(lr){5-7}\cmidrule(lr){8-10}
        & IoU $\uparrow$ & Chamfer $\downarrow$ & Track $\downarrow$ & IoU $\uparrow$ & Chamfer $\downarrow$ & Track $\downarrow$ & IoU $\uparrow$ & Chamfer $\downarrow$ & Track $\downarrow$ \\
        \midrule
        \textbf{DCA (ours)} & \textbf{0.721} & \textbf{0.50} & \textbf{1.01} & \textbf{0.748} & \textbf{1.30} & \textbf{3.19} & \textbf{0.756} & \textbf{1.19} & \textbf{2.02} \\
        rigid               & $0.675^{\dagger}$ & $0.51^{\dagger}$ & $1.14^{\dagger}$ & 0.718 & 1.65 & 3.72 & NaN & NaN & NaN \\
        single actuator     & 0.510 & 0.92 & 2.40 & 0.591 & 2.74 & 5.69 & 0.534 & 4.26 & 8.16 \\
        \bottomrule
    \end{tabular}
\end{table}

\begin{table}[H]
    \centering
    \caption{MoCE ablation: overall test metrics. Chamfer and Track in units of $10^{-2}$.}
    \label{tab:ablation_moe_overall}
    \small
    \setlength{\tabcolsep}{4pt}
    \begin{tabular}{l ccc ccc}
        \toprule
        \multirow{2}{*}{Variant} & \multicolumn{3}{c}{Future dynamics} & \multicolumn{3}{c}{Future appearance} \\
        \cmidrule(lr){2-4}\cmidrule(lr){5-7}
        & IoU $\uparrow$ & Chamfer $\downarrow$ & Track $\downarrow$ & PSNR $\uparrow$ & SSIM $\uparrow$ & LPIPS $\downarrow$ \\
        \midrule
        \textbf{MoCE (ours)} & \textbf{0.748} & \textbf{1.14} & \textbf{2.40} & \textbf{25.41} & \textbf{0.936} & \textbf{0.061} \\
        Single Neo-Hookean   & 0.658 & 1.94 & 3.75 & 23.84 & 0.928 & 0.077 \\
        \bottomrule
    \end{tabular}
\end{table}

\begin{table}[H]
    \centering
    \caption{Constitutive ablation: replacing the MoCE mixture with a single canonical Neo-Hookean expert. Per-category dynamics on the PhysTwin benchmark; Chamfer and Track in units of $10^{-2}$. (The MoCE ablation keeps the neural residual active. Since the residual can absorb part of the mismatch caused by a simplified constitutive model, this comparison may underestimate the contribution of MoCE.)}
    \label{tab:ablation_moe}
    \small
    \setlength{\tabcolsep}{3.5pt}
    \resizebox{\linewidth}{!}{%
    \begin{tabular}{l ccc ccc ccc}
        \toprule
        \multirow{2}{*}{Variant} & \multicolumn{3}{c}{Linear ($n{=}3$)} & \multicolumn{3}{c}{Planar ($n{=}9$)} & \multicolumn{3}{c}{Volumetric ($n{=}8$)} \\
        \cmidrule(lr){2-4}\cmidrule(lr){5-7}\cmidrule(lr){8-10}
        & IoU $\uparrow$ & Chamfer $\downarrow$ & Track $\downarrow$ & IoU $\uparrow$ & Chamfer $\downarrow$ & Track $\downarrow$ & IoU $\uparrow$ & Chamfer $\downarrow$ & Track $\downarrow$ \\
        \midrule
        \textbf{MoCE (ours)} & \textbf{0.721} & \textbf{0.50} & \textbf{1.01} & \textbf{0.748} & \textbf{1.30} & \textbf{3.19} & \textbf{0.756} & \textbf{1.19} & \textbf{2.02} \\
        Single Neo-Hookean   & 0.642 & 0.69 & 1.48 & 0.693 & 2.04 & 4.65 & 0.617 & 2.30 & 3.59 \\
        \bottomrule
    \end{tabular}%
    }
\end{table}

\begin{table}[H]
    \centering
    \caption{Stage~3 RGB-guided refinement: overall test metrics. Chamfer and Track in units of $10^{-2}$.}
    \label{tab:stage3_ablation_overall}
    \small
    \setlength{\tabcolsep}{4pt}
    \begin{tabular}{l ccc ccc}
        \toprule
        \multirow{2}{*}{Variant} & \multicolumn{3}{c}{Future dynamics} & \multicolumn{3}{c}{Future appearance} \\
        \cmidrule(lr){2-4}\cmidrule(lr){5-7}
        & IoU $\uparrow$ & Chamfer $\downarrow$ & Track $\downarrow$ & PSNR $\uparrow$ & SSIM $\uparrow$ & LPIPS $\downarrow$ \\
        \midrule
        Dynamics loss only  & 0.745 & 1.16 & 2.43 & 25.32 & 0.935 & 0.062 \\
        RGB-guided refinement (ours) & \textbf{0.748} & \textbf{1.14} & \textbf{2.40} & \textbf{25.41} & \textbf{0.936} & \textbf{0.061} \\
        \bottomrule
    \end{tabular}
\end{table}

\begin{table}[H]
    \centering
    \caption{Ablation of RGB-guided dynamics refinement in Stage~3. Per-category dynamics on the PhysTwin benchmark; Chamfer and Track in units of $10^{-2}$.}
    \label{tab:stage3_ablation}
    \small
    \setlength{\tabcolsep}{3.5pt}
    \resizebox{\linewidth}{!}{%
    \begin{tabular}{l ccc ccc ccc}
        \toprule
        \multirow{2}{*}{Variant} & \multicolumn{3}{c}{Linear ($n{=}3$)} & \multicolumn{3}{c}{Planar ($n{=}9$)} & \multicolumn{3}{c}{Volumetric ($n{=}8$)} \\
        \cmidrule(lr){2-4}\cmidrule(lr){5-7}\cmidrule(lr){8-10}
        & IoU $\uparrow$ & Chamfer $\downarrow$ & Track $\downarrow$ & IoU $\uparrow$ & Chamfer $\downarrow$ & Track $\downarrow$ & IoU $\uparrow$ & Chamfer $\downarrow$ & Track $\downarrow$ \\
        \midrule
        Dynamics loss only       & \textbf{0.725} & \textbf{0.50} & \textbf{1.01} & 0.743 & 1.33 & 3.22 & 0.752 & 1.22 & 2.08 \\
        RGB-guided refine        & 0.721 & \textbf{0.50} & \textbf{1.01} & \textbf{0.748} & \textbf{1.30} & \textbf{3.19} & \textbf{0.756} & \textbf{1.19} & \textbf{2.02} \\
        \bottomrule
    \end{tabular}%
    }
\end{table}

\begin{table}[H]
    \centering
    \caption{Online inference breakdown with 22,132 physical particles, 102,613 Gaussians, a $32^3$ grid, 42 MPM substeps per frame, and $848\times480$ rendering. The derived end-to-end latency is the sum of independently timed dynamics and appearance components.}
    \label{tab:runtime_breakdown}
    \small
    \begin{tabular}{lcc}
        \toprule
        Component & Mean latency (ms) & Share \\
        \midrule
        Dynamics: MPM substeps & 43.026 & 56.90\% \\
        Dynamics: Neural residual & 19.359 & 25.60\% \\
        Other dynamics overhead & 0.289 & 0.38\% \\
        Particle-to-Gaussian deformation & 11.289 & 14.93\% \\
        Gaussian rendering & 1.649 & 2.18\% \\
        \midrule
        Derived end-to-end & 75.612 & 100\% \\
        \bottomrule
    \end{tabular}
\end{table}

\begin{table}[H]
    \centering
    \caption{Dynamics-only runtime and peak GPU memory scaling on ``double\_\allowbreak{}lift\_\allowbreak{}sloth''; medians over nine measured repeats. Bold entries denote the default configuration. Each sweep varies only the indicated factor while holding the others fixed.}
    \label{tab:runtime_scaling}
    \small
    \setlength{\tabcolsep}{3pt}
    \begin{tabular}{lccc}
        \toprule
        Sweep & Setting & Median latency & GPU memory (GB) \\
        \midrule
        \multirow{3}{*}{Particles} & 6,660 & 50.267\,ms/frame & $\approx$1.1 \\
                 & 9,133 & 48.656\,ms/frame & $\approx$1.3 \\
                 & \textbf{22,132} & \textbf{62.443}\,ms/frame & $\boldsymbol{\approx}$\textbf{2.3} \\
        \midrule
        \multirow{3}{*}{Grid resolution} & $\mathbf{32}$ & \textbf{63.303}\,ms/frame & $\approx$2.3 \\
                        & 48 & 61.212\,ms/frame & $\approx$2.3 \\
                        & 64 & 59.888\,ms/frame & $\approx$2.3 \\
        \midrule
        \multirow{3}{*}{Substeps} & 21 & 42.184\,ms/frame & $\approx$2.3 \\
                 & \textbf{42} & \textbf{63.303}\,ms/frame & $\approx$2.3 \\
                 & 84 & 111.333\,ms/frame & $\approx$2.3 \\
        \midrule
        \multirow{3}{*}{Rollout horizon} & 10 frames & 0.580\,s/rollout & $\approx$2.3 \\
                        & 30 frames & 1.852\,s/rollout & $\approx$2.3 \\
                        & \textbf{50 frames} & \textbf{3.165}\,s/rollout & $\approx$2.3 \\
        \bottomrule
    \end{tabular}
\end{table}

\begin{table}[H]
    \centering
    \caption{Preprocessing and one-time training costs.}
    \label{tab:offline_costs}
    \small
    \setlength{\tabcolsep}{4pt}
    \begin{tabular}{lcc}
        \toprule
        Stage & Time & Peak GPU memory \\
        \midrule
        Preprocessing: depth/camera/geometry & 1 min 45.85 s & 19.45\,GB \\
        Preprocessing: segmentation/point tracking & 6 min 22.29 s & 10.94\,GB \\
        Training stage: dynamics learning & $\approx$2.5 h & $\approx$20\,GB \\
        Training stage: DCA gain selection & $\approx$16.6 min/scene & $\approx$2.30\,GB \\
        Training stage: appearance learning + RGB-guided refinement & $\approx$2--3 h & $\approx$20\,GB \\
        \bottomrule
    \end{tabular}
\end{table}

\end{document}